\documentclass[10pt,journal,compsoc]{IEEEtran}

\usepackage{amsmath,amsfonts}
\usepackage{algorithmic}
\usepackage{algorithm}
\usepackage{array}
\usepackage[caption=false,font=normalsize,labelfont=sf,textfont=sf]{subfig}
\usepackage{textcomp}
\usepackage{stfloats}
\usepackage{url}
\usepackage{verbatim}
\usepackage{graphicx}
\usepackage{cite}
\usepackage{multirow}
\usepackage{threeparttable}
\usepackage{tikz}
\usepackage[edges]{forest}
\usepackage{bbm}
\usepackage{pbox}
\usepackage{bm}
\usepackage{colortbl}
\usepackage[pagebackref,breaklinks,colorlinks]{hyperref}
\usepackage{fancyhdr}

\usepackage[capitalize]{cleveref}
\crefname{section}{Sec.}{Secs.}
\Crefname{section}{Section}{Sections}
\Crefname{table}{Table}{Tables}
\crefname{table}{Tab.}{Tabs.}
\crefname{equation}{Eq.}{Eqs.}

\definecolor{yellow}{rgb}{1, 1, 0.7}
\definecolor{orange}{rgb}{1, 0.85, 0.7}
\definecolor{cred}{rgb}{1, 0.7, 0.7}

\DeclareMathOperator*{\argmax}{arg\,max}
\newcommand\customparagraph[1]{\vspace{0.4em}\noindent\textbf{#1.}}

\def\eg{\emph{e.g.}}
\def\ie{\emph{i.e.}}

\ifCLASSINFOpdf
\else
\fi

\fancypagestyle{firstpage}{
  \fancyhf{}
  \fancyfoot[C]{\footnotesize © 2025 IEEE.  Personal use of this material is permitted.  Permission from IEEE must be obtained for all other uses, in any current or future media, including reprinting/republishing this material for advertising or promotional purposes, creating new collective works, for resale or redistribution to servers or lists, or reuse of any copyrighted component of this work in other works.
  }

}

\begin{document}
\title{Lightweight and Accurate Multi-View Stereo with Confidence-Aware Diffusion Model}

\author{Fangjinhua Wang\textsuperscript{*}, Qingshan Xu\textsuperscript{*}, Yew-Soon Ong\textsuperscript{\dag},~\IEEEmembership{Fellow,~IEEE},  Marc Pollefeys\textsuperscript{\dag},~\IEEEmembership{Fellow,~IEEE}%
\IEEEcompsocitemizethanks{
\IEEEcompsocthanksitem Fangjinhua Wang and Marc Pollefeys are with the Department of
Computer Science, ETH Zurich, Switzerland.
\IEEEcompsocthanksitem Qingshan Xu and Yew-Soon Ong are with the College of Computing and Data Science, Nanyang Technological University, Singapore.
\IEEEcompsocthanksitem Yew-Soon Ong is also with Center for Frontier AI Research, Institute of High Performance Computing, A*STAR, Singapore.
\IEEEcompsocthanksitem Marc Pollefeys is additionally with Microsoft, Zurich.
\IEEEcompsocthanksitem This work is partially supported by a research grant from Microsoft and by the National Research Foundation (NRF), Singapore, through the AI Singapore Programme under the project titled ``AI-based Urban Cooling Technology Development'' (Award No. AISG3-TC-2024-014-SGKR). 
}%

\thanks{\textsuperscript{*}: Equal contribution. \textsuperscript{\dag}: Corresponding authors.}
\thanks{Digital Object Identifier: 10.1109/TPAMI.2025.3597148}
}

\markboth{Journal of \LaTeX\ Class Files,~Vol.~14, No.~8, August~2015}%
{Shell \MakeLowercase{\textit{et al.}}: Bare Demo of IEEEtran.cls for Computer Society Journals}

\IEEEtitleabstractindextext{%
\begin{abstract}
To reconstruct the 3D geometry from calibrated images, learning-based multi-view stereo (MVS) methods typically perform multi-view depth estimation and then fuse depth maps into a mesh or point cloud. To improve the computational efficiency, many methods initialize a coarse depth map and then gradually refine it in higher resolutions. Recently, diffusion models achieve great success in generation tasks. Starting from a random noise, diffusion models gradually recover the sample with an iterative denoising process. In this paper, we propose a novel MVS framework, which introduces diffusion models in MVS. Specifically, we formulate depth refinement as a conditional diffusion process. Considering the discriminative characteristic of depth estimation, we design a condition encoder to guide the diffusion process. To improve efficiency, we propose a novel diffusion network combining lightweight 2D U-Net and convolutional GRU. Moreover, we propose a novel confidence-based sampling strategy to adaptively sample depth hypotheses based on the confidence estimated by diffusion model. Based on our novel MVS framework, we propose two novel MVS methods, DiffMVS and CasDiffMVS. DiffMVS achieves competitive performance with state-of-the-art efficiency in run-time and GPU memory. CasDiffMVS achieves state-of-the-art performance on DTU, Tanks \& Temples and ETH3D. Code will be available at: \url{https://github.com/cvg/diffmvs}.
\end{abstract}

}

\maketitle
\thispagestyle{firstpage}
\IEEEdisplaynontitleabstractindextext

\IEEEpeerreviewmaketitle

\section{Introduction}

\IEEEPARstart{M}{ulti-view} Stereo (MVS) aims to reconstruct the dense 3D geometry for an observed scene from a set of calibrated images. It has wide applications in real-world scenarios, such as robotics, %
autonomous driving, virtual / mixed reality and ``metaverse''. It typically performs multi-view depth estimation and then fuses depth maps into a point cloud or mesh~\cite{furukawa2015multi,schoenberger2016colmap,yao_2018_mvsnet,xu_2019_acmm}. When estimating depth maps, MVS is essentially an optimal correspondence search problem in a finite continuous depth space with photometric consistency assumption~\cite{schoenberger2016colmap}. However, due to the interference from illumination changes, non-Lambertian surfaces and low-textured areas which are common in real-world scenes, it is challenging to accurately estimate the depth.

\begin{figure}[tbp]
\centering
{\includegraphics[width=0.9\columnwidth]{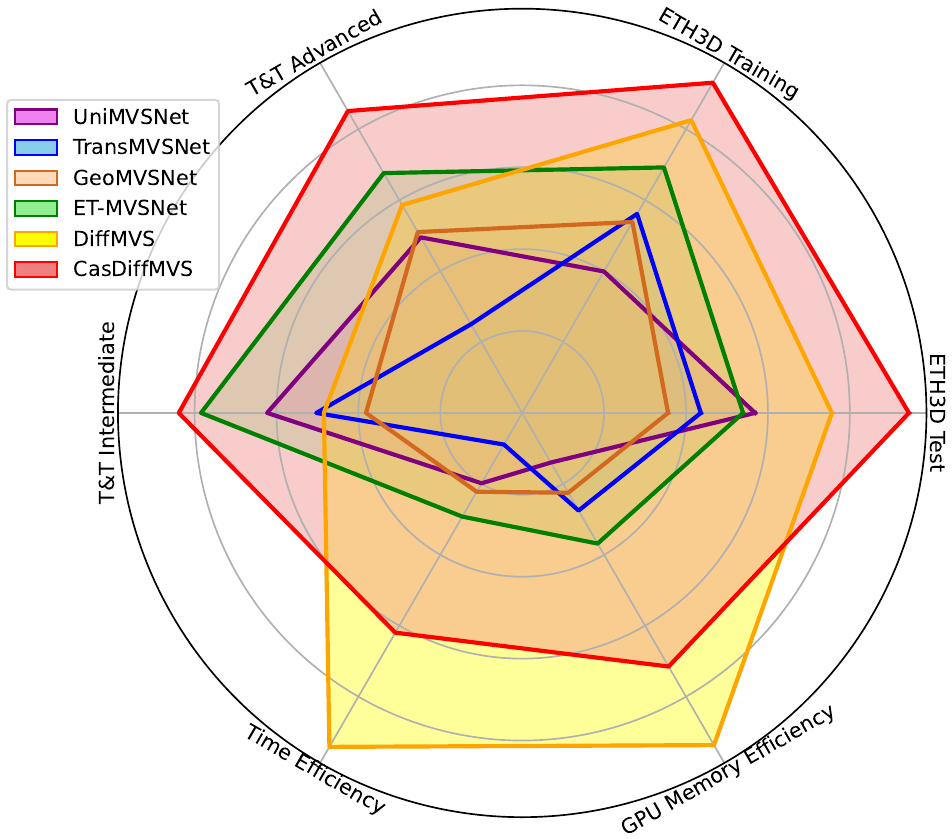}}
\caption{
\textbf{Comparison between our methods (DiffMVS and CasDiffMVS) and state-of-the-art learning-based MVS methods on reconstruction performance, efficiency in run-time and GPU memory}. While being more efficient in terms of run-time and GPU memory, CasDiffMVS achieves better reconstruction performance than UniMVSNet~\cite{peng2022rethinking}, TransMVSNet~\cite{ding2022transmvsnet}, GeoMVSNet~\cite{zhang2023geomvsnet} and ET-MVSNet~\cite{liu2023epipolar} on Tanks \& Temples (T\&T)~\cite{knapitsch2017tanks} and ETH3D~\cite{2017eth3d}. Moreover, DiffMVS achieves highest efficiency in run-time and GPU memory, while achieving competitive performance on T\&T and ETH3D. Note that for fair comparison, we evaluate the efficiency of all methods with the same input images on one workstation. 
}
\label{fig:teaser}
\end{figure}

To address this problem, traditional methods design many hand-crafted patch similarity metrics and use the plane sweep algorithm~\cite{planesweeping} to determine the optimal depth in discrete search space~\cite{merrell2007real}. However, their discrete solution usually cannot find the optima~\cite{campbell2008using}. PatchMatch MVS methods~\cite{galliani_2015_gipuma,schoenberger2016colmap,xu_2019_acmm} leverage the idea of nearest neighbor search and random perturbations to progressively search the optima, which improves the precision of depth estimation. However, they still struggle under the aforementioned challenges due to the traditional hand-crafted modeling. 

Recently, learning-based MVS methods demonstrate a significant improvement in reconstruction quality with deep image features and regularization learned by Convolutional Neural Networks (CNN). Typically, these methods~\cite{yao_2018_mvsnet,xu2020learning_inverse} use the plane sweep algorithm~\cite{planesweeping} to warp source image features into the reference view and then construct 3D cost volumes, which are regularized by 3D CNN for depth estimation. However, the precision of these methods are still constrained by the limited discrete depth ranges. To alleviate this, recurrent methods~\cite{yao_2019_rmvsnet,aarmvsnet,xu2021non} are proposed to increase the number of sampling depth. But these methods significantly sacrifice the efficiency due to their sequential processing for the 3D cost volumes. On the other hand, coarse-to-fine methods~\cite{gu_2020_cascademvsnet,cheng_2020_ucsnet,yang_2020_cvpr,patchmatchnet_wang} initialize a low-resolution depth map with sparse samples in the depth range. Then they gradually refine it in higher resolutions with a reduced finer depth range. Therefore, these methods further improve depth map accuracy. However, their refinement heavily relies on the quality of the coarse depth map since the estimation in higher resolutions is usually restricted in a reduced depth range that is centered around the coarse depth. If the coarse estimation is incorrect, it is difficult to recover from the errors induced at coarse resolutions and the estimation may be restricted in local minima.

In this work, we rethink how to perform %
accurate multi-view depth estimation by not only conditioning on the previous coarse estimation, but also by introducing random perturbations. Recently, diffusion models~\cite{ho2020denoising,rombach2022high} show that injecting random noise can avoid collapsing into local minima~\cite{song2020score}. Inspired by the diffusion models that can recover data samples from random noise with iterative denoising process~\cite{ho2020denoising,rombach2022high}, we hope our multi-view depth estimation can mimic the denoising process to introduce random noise and produce accurate estimation. However, since the multi-view depth estimation is a discriminative perception task instead of unconditional generative task, introducing diffusion-based denoising process into multi-view depth estimation will face the following challenges:
\begin{itemize}
    \item \textbf{Diffusion conditions}. In order to obtain accurate deterministic estimation, it is important to condition the diffusion network on some specific guidance. Since matching information is crucial to obtain accurate depth estimation in MVS, it is naturally to introduce matching information as diffusion condition~\cite{shao2022diffustereo}. However, whether MVS requires other diffusion conditions or how to integrate matching information into diffusion models is still under-explored. 
    \item \textbf{Diffusion sampling}. For generative tasks, diffusion models~\cite{ho2020denoising,rombach2022high} typically denoise a noisy sample with the information of this sample only.  %
    For discriminative tasks,~\eg, feature matching, using the information of a single sample only provides zero-order optimization information, which makes it unable to leverage non-local information for more accurate first-order optimization~\cite{xu_2019_acmm,teed2020raft,wang2022itermvs,ren2023hierarchical}. This hinders accurate estimation for discriminative perception tasks with diffusion models.   
    \item \textbf{Diffusion efficiency}. Classical diffusion models for generation tasks~\cite{ho2020denoising,rombach2022high} usually adopt large U-Nets with attention mechanisms~\cite{vaswani2017attention} to achieve impressive performance. In addition, some recent works~\cite{chen2022diffusiondet} have shown that stacking U-Nets is beneficial to improve performance. However, these designs will inevitably hinder the efficiency, which is important for MVS applications. 
\end{itemize}

To overcome the above challenges, we present a novel MVS framework %
with conditional diffusion models. Specifically, based on the coarse depth initialization, we design three key modules to construct our conditional diffusion models to refine the coarse depth map. 
First, we propose a condition encoder to effectively adapt diffusion models for depth estimation tasks. The condition encoder fuses matching information, image context and depth context features as the condition feature. This enables our diffusion model to perceive not only local similarity indicated by matching information, but also long-range context information provided by image and depth context features. 
Second, we introduce a confidence-based sampling strategy to adaptively generate multiple depth hypotheses for each pixel. %
Different from existing MVS methods that sample depth hypotheses around coarse estimation~\cite{gu_2020_cascademvsnet,patchmatchnet_wang}, our sampling is based on the \textit{noisy} coarse estimation perturbed by the random noise from diffusion process. This introduces randomness but may lack informative first-order optimization information if we generate depth hypotheses in a \textit{fixed} range~\cite{gu_2020_cascademvsnet,patchmatchnet_wang} since the coarse estimation may be perturbed a lot by random noise. To solve this, our confidence-based sampling strategy leverages per-pixel predicted confidence to adaptively adjust the sampling range and capture non-local first-order optimization information, thus facilitating the denoising process. 
Third, we propose a novel lightweight diffusion network, which consists of 2D denoising U-Net and convolutional GRU. %
Since GRU can capture historical information iteratively to enhance feature expression capability and is proven effective in iterative refinement~\cite{teed2020raft}, we design a lightweight U-Net combined with convolutional GRU as our diffusion model. We perform multi-iteration refinement with  GRU in a single diffusion timestep, which not only improves performance but also circumvents the usage of large or stacked U-Nets of existing diffusion models~\cite{ho2020denoising,chen2022diffusiondet} and thus improves efficiency. %

With our framework, we propose two novel MVS methods, named as DiffMVS and CasDiffMVS. DiffMVS performs depth refinement with a single-stage diffusion model, while CasDiffMVS extends DiffMVS and performs depth refinement on two stages. The former is tailored for real-time applications, while the latter is designed for high-accuracy requirements. %
Extensive experiments demonstrate that DiffMVS achieves competitive performance with state-of-the-art (SOTA) efficiency in both run-time and memory, while CasDiffMVS achieves SOTA reconstruction performance on various benchmarks with high efficiency, as shown in Fig.~\ref{fig:teaser}. In summary, our contributions are as follows:
\begin{itemize}
    \item We propose a novel MVS framework which exploits conditional diffusion models to achieve efficient, accurate and lightweight multi-view depth estimation. 
    \item We propose a condition encoder to fuse matching information, image context and depth context features as the condition input of diffusion model. This provides the diffusion model specific guidance to generate accurate depth predictions. 
    \item We introduce a confidence-based sampling strategy, which adaptively generates multiple samples in a local range based on the confidence estimated from diffusion model to provide informative first-order optimization directions. %
    \item We develop a lightweight diffusion network, which leverages convolutional GRU to replace the large denoising U-Nets used in classical diffusion models. This greatly improves the efficiency of our models. 
    \item Based on our novel MVS framework, we propose two novel MVS methods, DiffMVS and CasDiffMVS. The former one achieves competitive performance with SOTA efficiency in both run-time and memory, while the later one achieves SOTA performance on various benchmarks. 
\end{itemize}

\section{Related Work}
\customparagraph{Traditional MVS}
Traditional MVS methods can be mainly divided into three categories: 
voxel based~\cite{ulusoy2017, kutulakos2000theory, seitz1999photorealistic, kostrikov2014probabilistic}, point cloud based~\cite{lhuillier2005quasi, furukawa2010} and depth map based~\cite{galliani_2015_gipuma, schonberger2016pixelwise, xu_2019_acmm,xu2020acmp,xu2022multi}. 
Depth map based methods decouple the reconstruction problem into multi-view depth estimation and depth fusion, which explicitly improve flexibility and scalability. This characteristic makes depth map based methods dominate MVS. 
PatchMatch~\cite{barnes2009patchmatch} algorithm is commonly used by traditional depth map based methods to improve efficiency. 
For example, Gipuma~\cite{galliani_2015_gipuma} uses a
red-black checkerboard pattern to propagate depth hypotheses. ACMM~\cite{xu_2019_acmm} further adopts adaptive checkerboard
sampling and multi-scale geometric consistency guidance to improve performance. To further alleviate matching ambiguity in low-textured areas, HPM-MVS~\cite{ren2023hierarchical} introduces hierarchical prior mining for non-local MVS. 
ADP-MVS~\cite{wang2023adaptive} designs adaptive patch deformation to measure matching cost. 
These PatchMatch MVS methods leverage random neighbor search and perturbations to estimate depth from a finite continuous depth space. However, 
traditional MVS methods rely on hand-crafted matching metrics and thus encounter challenges in challenging conditions,~\eg, illumination changes, low-textured areas, and non-Lambertian surfaces~\cite{aanaes2016_dtu,knapitsch2017tanks,2017eth3d,yao2020blendedmvs}. Our framework leverages deep features to measure matching similarities and introduces randomness from diffusion models to avoid local minima. 

\customparagraph{Learning-based MVS}
Due to the limited performance of hand-crafted modeling under challenging conditions, many learning-based MVS methods have been proposed in recent years and achieved better performance on various benchmarks~\cite{aanaes2016_dtu,knapitsch2017tanks,2017eth3d}. 
Based on plane sweep algorithm~\cite{collins1996space}, MVSNet~\cite{yao_2018_mvsnet} proposes differentiable homography to construct a 3D cost volume with warped image features, regularizes the cost volume with a 3D U-Net~\cite{ronneberger2015u}, and regresses the depth map from the probability volume. 
However, 3D CNN explicitly increases the memory consumption and run-time. 
To improve computational efficiency, many variants based on MVSNet~\cite{yao_2018_mvsnet} are proposed and can be mainly categorized into recurrent methods~\cite{yao_2019_rmvsnet, d2hcrmvsnet,aarmvsnet} and coarse-to-fine methods~\cite{gu_2020_cascademvsnet, cheng_2020_ucsnet, yang_2020_cvpr, xu2020pvsnetpv, vismvsnet, patchmatchnet_wang, ma2021epp, ding2022transmvsnet, peng2022rethinking, xu2022learning, wang2022mvster}. 
Recurrent methods reduce the memory consumption by sequentially regularizing 2D cost maps from the 3D cost volume with recurrent neural networks (RNN). However, these methods have low time efficiency since they trade time for memory. 
Coarse-to-fine methods first estimate a coarse depth map in low resolution and then reduce the search range in higher resolutions to improve the accuracy. These methods explicitly improve the inference speed and reduce memory consumption. 
Recently, inspired by RAFT~\cite{teed2020raft} that uses a GRU~\cite{cho2014learning} to emulate first-order optimization, several methods~\cite{wang2022itermvs,wang2022efficient} use GRU for depth refinement and achieve high efficiency in memory and run-time. Different from these methods, our methods leverage the denoising process of diffusion models to refine the coarse depth estimation.

\customparagraph{Geometry estimation with diffusion models}
As a class of deep generative models, diffusion models~\cite{ho2020denoising,song2019generative, song2020score} start from a random noise and recover the data sample with an iterative denoising process. 
They have achieved impressive results in image~\cite{ho2020denoising,dhariwal2021diffusion,rombach2022high} and video~\cite{ho2022video} generation tasks. 
Recently, many researchers have shown that diffusion models can be used for many geometry estimation tasks, such as monocular depth estimation~\cite{duan2023diffusiondepth,ji2023ddp,ke2024repurposing}, monocular normal estimation~\cite{fu2024geowizard,ye2024stablenormal}, depth completion~\cite{guo2025murre}, feature matching~\cite{nam2023diffusion}, pose estimation~\cite{wang2023posediffusion,zhang2024cameras,lu2025matrix3d} and optical flow estimation~\cite{saxena2023surprising, dong2023open}. 
With the introduction of conditional diffusion models, these methods outperform previous state-of-the-art methods. In this paper, we introduce a novel conditional diffusion model in MVS and explore a generative paradigm for accurate and efficient reconstruction.

\customparagraph{Confidence estimation in stereo/MVS}
Confidence estimation has been shown to be effective in predicting the reliability of disparity/depth estimation in stereo/MVS. Early methods~\cite{park2018learning,poggi2016learning,haeusler2013ensemble} utilize conventional features to train random forest classifiers for confidence estimation with a two-class label for each pixel. In stereo matching, recent methods~\cite{tosi2018beyond,shaked2017improved,kim2019laf} leverage CNN to predict confidence from disparity map, reference image and cost volume. 
In MVS, some methods also use confidence/uncertainty estimation to improve depth prediction. UCSNet~\cite{cheng_2020_ucsnet} computes the variance of the probability volume to define the confidence and uses it to determine the depth range at the next stage. UGNet~\cite{su2022uncertainty} employs uncertainty learning to predict the uncertainty from the cost volume directly
to capture the uncertainty of the depth map. Vis-MVSNet~\cite{vismvsnet} estimates the uncertainty of the pair-wise depth map to construct a more reliable aggregated cost volume to predict the final depth map. Our methods predict confidence based on our designed diffusion model, which incorporates cost volume, image context and depth context features. Moreover, our predicted confidence is combined with the noisy depth for depth sampling, which greatly enhances the power of our diffusion models. 

\section{Methodology}

\begin{figure*}[tbp]
\centering
{\includegraphics[width=\linewidth]{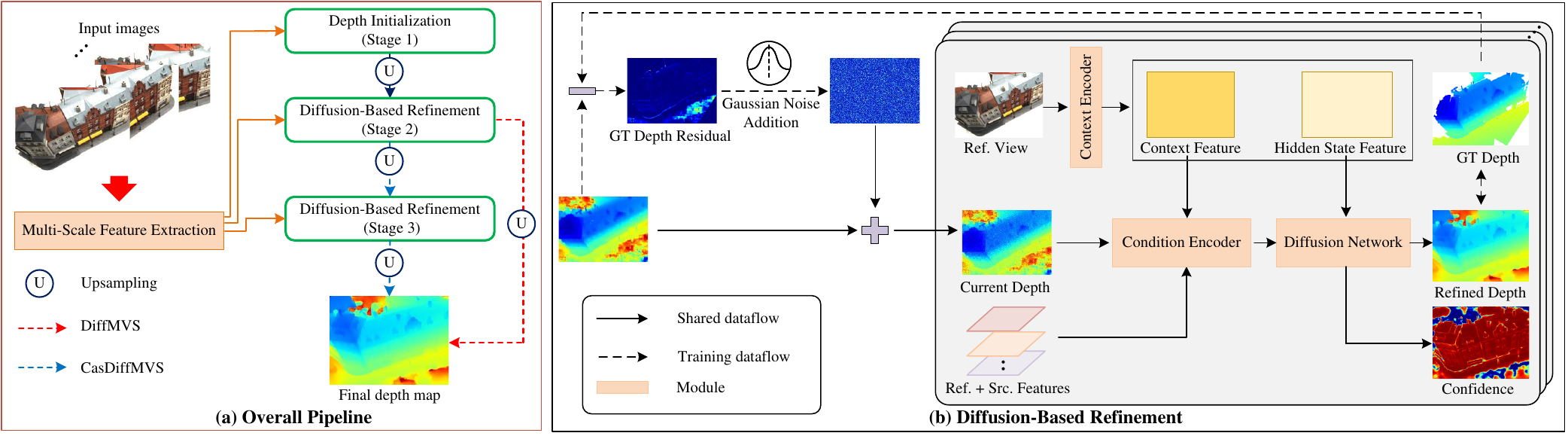}}
\caption{
\textbf{Overview of DiffMVS and CasDiffMVS.} 
(a) shows the overall pipeline of DiffMVS and CasDiffMVS. A coarse depth map is initialized in low resolution (stage 1). Then DiffMVS uses a single diffusion-based refinement module on stage 2 to refine the coarse depth map, while CasDiffMVS uses cascade diffusion-based refinement on stage 2, 3. Finally, the refined depth map is upsampled to full resolution. 
(b) shows the architecture of diffusion-based refinement. It refines the coarse depth map by denoising the current \textit{noisy} depth map. A context encoder is used to extract image context and hidden state features from the reference view. 
To encode condition feature for the diffusion network, a condition encoder takes as input the image context feature, depth context feature and the cost volume constructed with reference and source features. With the condition feature and hidden state feature, the diffusion network outputs refined depth and confidence. 
Best viewed on a screen when zoomed in. 
}\label{fig:overview}
\end{figure*}

In this section, we introduce the details of our MVS framework with conditional diffusion models. The framework mainly consists of two modules: depth initialization and diffusion-based depth refinement. First, we obtain an initial coarse depth map with the depth initialization module. Then we develop a novel conditional diffusion model to refine it in higher resolutions. Based on our MVS framework, we propose DiffMVS and CasDiffMVS, which are visualized in Fig.~\ref{fig:overview}. Structually, DiffMVS performs depth refinement on a single stage (stage 2) and then upsamples to full resolution, while CasDiffMVS performs refinement on two stages (stage 2, 3) and then upsamples to full resolution.

\subsection{Preliminaries} 
As a class of generative models, diffusion models~\cite{ho2020denoising,song2019generative,song2020score} iteratively recover samples from random noise. %
Given the data distribution $\bm{x} \sim p(\bm{x})$, diffusion model defines a Markovian chain of \textit{forward} process and gradually adds Gaussian noise $\mathcal{N} (0, \bm{I})$ to the sample $\bm{x}_{0}$. 
The noisy sample $\bm{x}_{t}$ at timestep $t\in\{1,\dots,T\}$ can be computed as: 
\begin{equation}
\label{eq:forward_process}
    \bm{x}_t = \sqrt{\bar{\alpha}_t} \bm{x}_0 + \sqrt{1 - \bar{\alpha}_t} \bm{\epsilon}, \quad \bm{\epsilon} \sim \mathcal{N} (0, \bm{I}),
\end{equation}
where $\bar{\alpha}_t := \prod_{i=1}^{t} \alpha_i = \prod_{i=1}^{t} (1 - \beta_i)$ and $\beta_i$ represents the noise variance schedule~\cite{ho2020denoising}. 

In the \textit{reverse} process, the denoising model $f_\theta(\bm{x}_t, t)$, parameterized with learnable parameters $\theta$, is trained to predict the added noise $\hat{\bm{\epsilon}}$ %
from $\bm{x}_{t}$ by minimizing the training objective function $\mathcal{L}$ as follows~\cite{ho2020denoising}:
\begin{equation}
    \mathcal{L} = \mathbb{E}_{\bm{x}_0, \bm{\epsilon} \sim \mathcal{N} (0, \bm{I}), t \sim \mathcal{U} (1, T)} ||\bm{\epsilon} - \hat{\bm{\epsilon}}||^2, 
\end{equation}
where $\mathcal{U}(1, T)$ denotes the uniform distribution. To better control the generation, conditional diffusion models introduce condition $c$, \eg, text prompt~\cite{rombach2022high}, image~\cite{saxena2023monocular, ji2023ddp}, into the denoising model as $f_\theta(\bm{x}_t, c, t)$. Note that the diffusion models for discriminative tasks~\cite{karazija2023diffusion,chen2022diffusiondet,ji2023ddp,nam2023diffusion,liu2023difflow3d} are all conditional diffusion models. 

At the inference stage, $\bm{x}_{0}$ is reconstructed from a Gaussian noise $\bm{x}_{T}$ in an iterative way based on the diffusion network $f_\theta(\cdot)$ and an update rule~\cite{ho2020denoising,song2020denoising}. 
To achieve good performance, DDPM~\cite{ho2020denoising} needs to sample many steps and thus has low sampling efficiency. In contrast, DDIM~\cite{song2020denoising} proposes non-Markovian forward process, which is capable to generate high-quality samples with much fewer sampling steps.

\subsection{Problem Formulation and Motivation}

In this work, we aim to solve multi-view depth estimation tasks in MVS with a conditional diffusion model. Recently, most state-of-the-art MVS methods~\cite{gu_2020_cascademvsnet,patchmatchnet_wang,ding2022transmvsnet,wang2022efficient} initialize a coarse depth in low resolution and then refine it in high resolutions, which achieves impressive performance in both efficiency and accuracy. 
Following these methods, we initialize a low-resolution depth map and then progressively upsample and refine it with diffusion process in a cascade pipeline. 

Formally, for a reference image $\bm{I}_\text{0}$ and its neighboring images $\{\bm{I}_i\}_{i=1}^{N-1}$, we denote the image features extracted from these images as $\bm{F}_0$ and $\{\bm{F}_i\}_{i=1}^{N-1}$.  
The objective of multi-view depth estimation is to find the depth map $D$ corresponding to the reference image $\bm{I}_\text{0}$. 
In our framework, we aim to refine an initial depth map, $\bm{D}_{\text{init}}$, with a diffusion model and image features. Specifically, we seek to find $\bm{D}^*$ that maximizes the posterior probability of the estimated depth map given the initial depth map $\bm{D}_{\text{init}}$ and image features $\bm{\mathcal{F}}=\{\bm{F}_\text{0}, \{\bm{F}_i\}_{i=1}^{N-1}\}$,~\textit{i.e.}, $p(\bm{D}|\bm{\mathcal{F}}, \bm{D}_{\text{init}})$.  
To find the depth map $\bm{D}^*$ that maximizes the posterior, we can use the maximum a posteriori (MAP) approach: 
\begin{equation}\label{eq:map}
\begin{split}
    \bm{D}^* &= \argmax_{\bm{D}} p(\bm{D}|\bm{\mathcal{F}}, \bm{D}_{\text{init}}) = \argmax_{\bm{D}} p(\bm{\mathcal{F}}|\bm{D}) \cdot p(\bm{D}|\bm{D}_{\text{init}}) \\ &= \argmax_{\bm{D}} \{\log p(\bm{\mathcal{F}}|\bm{D}) + \log p(\bm{D}|\bm{D}_{\text{init}})\}.
\end{split}
\end{equation}
In this probabilistic interpretation, the first term (data term) represents the matching evidence between image features $\bm{F}_\text{0}$ and $\{\bm{F}_i\}_{i=1}^{N-1}$, and the second term (prior term) encodes prior knowledge of the depth map $\bm{D}$.

Previous methods focus on constructing cost volumes / matching confidence and regressing the final depth map, including TransMVSNet~\cite{ding2022transmvsnet}, MVSTER~\cite{wang2022mvster} and DUSt3R~\cite{wang2024dust3r}. This might be analogy to $\argmax_{\bm{D}} \log p(\bm{\mathcal{F}}|\bm{D})$. On the one hand, these methods usually construct the data term using a refined depth range centered around the initial depth map, $\bm{D}_{\text{init}}$. This often causes these methods to get stuck in local minima. On the other hand, the prior term $\log p(\bm{D}|\bm{D}_{\text{init}})$ is not explicitly considered in these regression-based methods. In fact, it is difficult for the data term to learn an accurate depth estimation in ambiguous areas,~\textit{e.g.} textureless areas and occlusions, which are common in the real world.

Since conditional diffusion models excel at learning the data distribution $p(\bm{D} | \bm{\mathcal{F}}, \bm{D}_{\text{init}})$ given inputs $\bm{\mathcal{F}}$ and $\bm{D}_{\text{init}}$, we can leverage them to explicitly incorporate both data term and prior term within a unified probabilistic framework. 
The data term encodes the geometry consistency between $\bm{F}_\text{0}$ and $\{\bm{F}_i\}_{i=1}^{N-1}$, which guides the generation process to be consistent with conditions. 
Meanwhile, the prior term encodes the structural and geometric constraints, which alleviates the ambiguity in challenging regions. Although diffusion models in image generation tasks enhance diversity without conditions, our model reduces diversity as it is conditioned on the geometry consistency between $\bm{F}_\text{0}$ and $\{\bm{F}_i\}_{i=1}^{N-1}$, which is a characteristic of conditional diffusion models~\cite{ho2022classifier}. The random noise within our diffusion model aims to avoid local minima. Therefore, inspired by the ability to avoid falling into the local minima and the explicit prior modeling in diffusion models, we aim to explore diffusion models for MVS.

\subsection{Multi-scale feature extraction}\label{sec::fea}
Given a reference image $\bm{I}_0 \in \mathbb{R}^{H \times W \times 3}$ and its neighboring source images $\{\bm{I}_i\}_{i=1}^{N-1} \in \mathbb{R}^{H \times W \times 3}$, we extract multi-scale features with a Feature Pyramid Network (FPN)~\cite{lin2017fpn}, where $N$ is the number of input images, $H$ and $W$ denote the image height and width, respectively. Specifically, we extract image features at $M$ different stages (DiffMVS: $M=2$, CasDiffMVS: $M=3$). 
We denote the image feature of $\bm{I}_i$ at stage $m$ with $\bm{F}_{i}^{m} \in \mathbb{R}^{H_m \times W_m \times C_m}$, where $H_m = H / 2^{4-m}, W_m = W / 2^{4-m}$, $m=1, \cdots, M$. 
For simplicity, we omit the superscript $m$ below. 

For the reference image $\bm{I}_0$, we further use a context FPN encoder to extract the multi-scale context features $\bm{F}_c$ and hidden state features $\bm{h}_0$ for upsampling and diffusion model.

\subsection{Depth initialization}\label{sec:di}
\begin{figure}[tbp]
\centering
{\includegraphics[width=\linewidth]{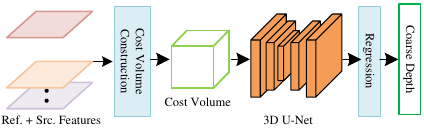}}
\caption{
\textbf{Pipeline of depth initialization.} A lightweight cost volume is constructed and regularized by a 3D U-Net to produce the initial depth $\bm{D}_{\text{init}}$. See Sec.~\ref{sec:di} for more details. Best viewed on a screen when zoomed in.
}\label{fig:di}
\end{figure}

We initialize the coarse depth map $\bm{D}_\text{init}$ at stage 1 ($1/8$ resolution) and then refine it in higher resolutions with diffusion model. The initialization pipeline is shown in Fig.~\ref{fig:di}. 

\customparagraph{Cost volume construction}
Given a pre-defined depth range $[d_{\text{min}}, d_{\text{max}}]$, we uniformly sample per-pixel depth hypotheses $\{d_j\}_{j=1}^{D_0}$ in the \textit{inverse} range $[1/{d_{\text{max}}}, 1/{d_{\text{min}}}]$, where $D_0$ is the number of initial samples. This inverse sampling is considered more suitable for large-scale scenes~\cite{xu2020learning_inverse,yao_2019_rmvsnet,patchmatchnet_wang,xu2022learning}. 
Then we warp the features of source views into the reference view at these depth hypotheses. 
Specifically, for a pixel $\bm{p}$ in the reference view, we compute the corresponding pixel $\bm{p}_{i,j}:=\bm{p}_i(d_j)$ in source view $i$ as follows: 
\begin{equation}
    \bm{p}_{i,j} = \bm{K}_i \cdot (\bm{R}_{0,i}\cdot(\bm{K}_0^{-1} \cdot \bm{p} \cdot d_j) + \bm{t}_{0,i}),
\end{equation}
where $\bm{K}_0, \bm{K}_i$ denote intrinsic and $[\bm{R}_{0,i}|\bm{t}_{0,i}]$ denotes relative transformation between reference view and source view $i$. 
Then we obtain the warped source feature $\bm{F}_i(\bm{p}_{i,j})$ with bilinear interpolation. 

To reduce computation, we follow~\cite{xu2020learning_inverse, patchmatchnet_wang, wang2022itermvs} and use group-wise correlation to compute the similarity between reference feature $\bm{F}_0(\bm{p})$ and warped source feature $\bm{F}_i(\bm{p}_{i,j})$.
By evenly dividing $C_1$ feature channels into $G\!=\!4$ groups, we compute the $g$-th group similarity 
$\bm{s}_{i}(\bm{p},j)^g$ as:
\begin{equation}\label{eq:similarity}
    \bm{s}_{i}(\bm{p}, j)^g = \frac{1}{C_1/G} \left \langle \bm{F}_0(\bm{p})^g, \bm{F}_i(\bm{p}_{i,j})^g\right \rangle,
\end{equation}
where $\left \langle \cdot, \cdot \right \rangle$ denotes the dot product. This results in $N-1$ similarity volumes $\{\bm{S}_{i}\}_{i=1}^{N-1}\in\mathbb{R}^{H_1 \!\times\! W_1 \!\times\! D_0 \!\times\! G}$.

To aggregate these $N-1$ similarity volumes, the na\"ive way is to compute the average of them~\cite{xu2020learning_inverse}. However, this does not take occlusions into consideration, which will lead to invalid matching and inaccurate estimations~\cite{schoenberger2016colmap}. 
Following~\cite{xu2020pvsnetpv,patchmatchnet_wang,xu2022learning}, we estimate the pixel-wise view weight for each source view $i$ with a lightweight network to provide visibility information and robustly aggregate matching information. 
Specifically, for each $\bm{S}_i$, we apply two 3D convolution layers and then \textit{softmax} along the depth dimension to produce $\bm{w}_i \in \mathbb{R}^{H_1 \!\times\! W_1 \!\times\! D_0}$. 
For pixel $\bm{p}$ and source view $i$, the corresponding view weight $\bm{W}_i(\bm{p})$ is computed as: 
\begin{equation}
    \bm{W}_i(\bm{p}) = \max_{j} \bm{w}_{i}(\bm{p},j).
\end{equation}
Finally, the cost volume $\bm{V}$ is aggregated as:
\begin{equation}\label{eq:cost_volume}
\setlength{\abovedisplayskip}{3pt}
\setlength{\belowdisplayskip}{3pt}
    \bm{V} = \frac{\sum_{i=1}^{N-1}\bm{W}_i \cdot \bm{S}_{i}}{\sum_{i=1}^{N-1}\bm{W}_i}.
\end{equation}

\customparagraph{Depth prediction} 
We regularize $\bm{V}$ with a lightweight 3D U-Net and then apply \textit{softmax} along the depth dimension to produce the probability volume $\bm{P} \in\mathbb{R}^{H_1 \!\times\! W_1 \!\times\! D_0}$. The initial depth $\bm{D}_{\text{init}}$ is computed as the expectation in \textit{inverse} range: 
\begin{equation}
    \bm{D}_{\text{init}} = \left(\sum_{j=1}^{D_0} \bm{P} (j) \cdot \frac{1}{d_{j}} \right)^{-1},
\end{equation}
where $\bm{P} (j)$ is the probability of all pixels at depth $d_j$.

\subsection{Diffusion-based refinement}\label{sec:dif-ref}
The initial depth $\bm{D}_{\text{init}}$ is not accurate since it is estimated at low resolution with sparse depth hypotheses. 
We propose a conditional diffusion model, shown in Fig.~\ref{fig:overview}(b), to refine the coarse depth on higher resolutions. 
Unlike unconditional diffusion models in generation tasks~\cite{ho2020denoising}, depth estimation is a discriminative perception task and thus our diffusion model is conditioned on the feature encoded by \textit{condition encoder}, so that the generation diversity can be constrained and thus depth map can be accurately denoised.

\customparagraph{Forward process}
To make our method robustly generalize to scenes with different scales, we use the normalized \textit{inverse} depth $\bm{\bar{D}} \in [0, 1]$ for depth map $\bm{D}$ throughout the diffusion process:
\begin{equation} \label{eq::normalization}
    \bm{\bar{D}} = \left( \frac{1}{\bm{D}} - \frac{1}{d_{\text{max}}} \right) / \left( \frac{1}{d_{\text{min}}} - \frac{1}{d_{\text{max}}} \right).
\end{equation}
For each refinement stage, we denote the initial depth upsampled from previous stage as $\bm{D}_0$. We compute the ground truth depth residual $\bm{x}_{0}$ as:
\begin{equation}
\label{eq:gt_residual}
    \bm{x}_{0} = \bm{\bar{D}}_{\text{gt}} - \bm{\bar{D}}_{0},
\end{equation}
where $\bm{\bar{D}}_{\text{gt}}$ denotes the normalized \textit{inverse} of ground truth. 
During training, we uniformly sample timestep $t \sim \bm{U}(1, T)$ and compute the noisy $\bm{x}_{t}$ with Eq.~\ref{eq:forward_process}.  %

\begin{figure}[tbp]
\centering
{\includegraphics[width=\linewidth]{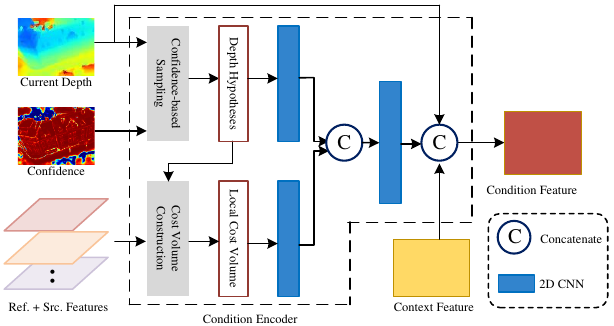}}
\caption{
\textbf{Structure of our condition encoder.} With the new depth hypotheses generated with confidence-based sampling, we compute the local cost volume as introduced in Sec.~\ref{sec:di}. Our condition encoder applies 2D convolution layers to encode geometric matching information from the local cost volume, depth hypotheses and image context feature as the condition feature. Best viewed on a screen when zoomed in.
}\label{fig:condition_encoder}
\end{figure}

\begin{figure}[tbp]
\centering
{\includegraphics[width=\linewidth]{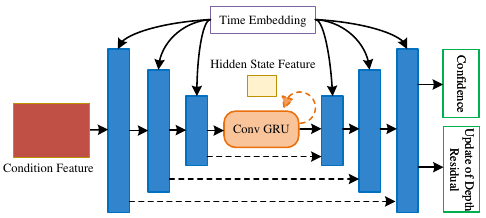}}
\caption{
\textbf{Structure of our diffusion network.} The network combines a 2D U-Net with convolutional GRU. The condition feature is used as the condition for 2D U-Net to denoise the depth residual. Best viewed on a screen when zoomed in.
}\label{fig:diffuison_net}
\end{figure}

\customparagraph{Reverse process}
To refine the depth, our diffusion model denoises the depth residual $\bm{x}_0$ with the reverse process. 
Following prior works~\cite{ho2020denoising,rombach2022high}, we design our conditional denoising diffusion model based on a 2D U-Net architecture, shown in Fig.~\ref{fig:diffuison_net}. 

Recently, RAFT~\cite{teed2020raft} is popular in optical flow estimation since it performs iterative refinements with GRU to achieve impressive performance with low complexity. 
Motivated by RAFT, many stereo~\cite{lipson2021raft} / MVS~\cite{wang2022itermvs,wang2022efficient,ma2022multiview} methods adopt iterative GRU refinements with lightweight 2D convolutions to outperform a single refinement with cumbersome 3D convolutions. 
Since we focus on both efficiency and performance in this work, we follow these methods and introduce convolutional GRU into a lightweight 2D U-Net (See below for details).  
Consequently, in each diffusion timestep, we iteratively refine $K$ times. 
Note that recent diffusion models,~\eg, DiffusionDet~\cite{chen2022diffusiondet} and DifFlow3D~\cite{liu2023difflow3d}, also update multiple iterations in each diffusion timestep to improve performance in discriminative perception tasks. 

Specifically, in the $k$-th iteration of diffusion timestep $t$, our diffusion model predicts the update of depth residual, $\Delta \bm{x}_{t,k}$, and the confidence $\bm{C}_{t,k}$ of current estimation.  %
After $K$ iterations, the depth residual is denoised as: 
\begin{equation}
    \hat{\bm{x}}_{0} = \bm{x}_{t} + \Delta \bm{x}_{t} = \bm{x}_{t} + \sum_{k=1}^K \Delta \bm{x}_{t,k},
\end{equation}
where $\hat{\bm{x}}_{0}$ is the denoised prediction of the ground truth residual $\bm{x}_{0}$. Based on Eq.~\ref{eq:gt_residual}, the final refined \textit{inverse} depth can be represented as $\bm{\bar{D}}_{t,K} = \bm{\bar{D}}_{0} + \hat{\bm{x}}_{0}$. 
In the $k$-th iteration, we compute the intermediate refined \textit{inverse} depth $\bm{\bar{D}}_{t,k}$ for training process: 
\begin{equation}
    \bm{\bar{D}}_{t,k} = \bm{\bar{D}}_{0} + \bm{x}_{t} + \sum_{n=1}^{k} \Delta \bm{x}_{t,n}.
\end{equation}

\customparagraph{Condition encoder}
For simplicity, we omit the subscript $t$ of diffusion timestep below. 
In the $k$-th iteration, we first sample $D_1$ per-pixel new hypotheses in a local range around previous depth estimation $\bm{\bar{D}}_{k-1}$, resulting in $\bm{\bar{D}}_{k}^{\text{sample}} \in\mathbb{R}^{{H}_{m} \!\times\! {W}_{m} \!\times\! D_{1}}$ (details of sampling strategy with confidence will be introduced later). 
Second, we compute the local cost volume $\bm{L}_{k} \in \mathbb{R}^{{H}_{m} \!\times\! {W}_{m} \!\times\! D_{1} \!\times\! G}$ for the new hypotheses (Sec.~\ref{sec:di}), which is further reshaped into $\mathbb{R}^{{H}_{m} \!\times\! {W}_{m} \!\times\! (D_{1} \!\times\! G)}$. For the view weights $\{\bm{W}_i\}_{i=1}^{N-1}$ in Eq.~\ref{eq:cost_volume}, we reuse those predicted at stage 1 and upsample them with nearest neighbors. 

For diffusion model, the depth samples as well as corresponding costs can guide the model to find a refined depth value. In addition, the consistency between depth map and image is also helpful to refine the depth map~\cite{patchmatchnet_wang}. 
Therefore, we propose a lightweight \textit{condition encoder} to inject useful conditions into the diffusion model. The structure of our condition encoder is depicted in Fig~\ref{fig:condition_encoder}. 
Specifically, the cost volume $\bm{L}_{k}$ is processed by 2D convolutional layers.  
Additionally, we apply 2D convolutional layers on $\bm{\bar{D}}_{k}^{\text{sample}}$ to generate depth context features. 
Then we concatenate these two features and apply 2D convolutional layers, which is finally concatenated with the previous depth $\bm{\bar{D}}_{k-1}$ and reference context feature $\bm{F}_c$ as the condition input of diffusion model.

\customparagraph{Denoising U-Net with GRU} 
Following prior works~\cite{ho2020denoising,rombach2022high}, we design our conditional diffusion model based on a 2D U-Net architecture. 
Following DDPM~\cite{ho2020denoising}, we inject the timestep embedding into the layers of U-Net. 
We experimentally find that the attention operation used in DDPM does not explicitly improve the performance. Therefore, we do not introduce attention into the U-Net so that the models are efficient and lightweight. 

As mentioned before, we introduce GRU into the 2D U-Net in consideration of efficiency and performance. 
The structure of our diffusion network is depicted in Fig.~\ref{fig:diffuison_net}. 
Specifically, we introduce a convolutional GRU in the lowest resolution of U-Net. %
To initialize the hidden state of GRU, we apply 2D convolutional layers followed by \textit{tanh} non-linearity on the hidden state feature $\bm{h}_0$ of the reference image. %
In the $k$-th iteration, the U-Net encodes the conditional feature from the \textit{condition encoder} to update the hidden state $\bm{h}_{k-1}$ of the GRU into $\bm{h}_k$. Then $\bm{h}_k$ is decoded to predict the update of depth residual $\Delta \bm{x}_{k}$, as well as the confidence $\bm{C}_{k}$ of current depth estimation (\textit{sigmoid} is applied on confidence to ensure $\bm{C}_{k} \in [0,1]$).

\customparagraph{Confidence-based sampling}
Utilizing non-local information with multiple samples is proven effective in many MVS methods to provide first-order optimization information~\cite{xu_2019_acmm,teed2020raft,wang2022itermvs,ren2023hierarchical}. 
Therefore, different from classical diffusion models~\cite{ho2020denoising} that use a single sample, we propose confidence-based sampling strategy and adaptively generate multiple samples. 
To refine the coarse depth map, many MVS methods~\cite{gu_2020_cascademvsnet,patchmatchnet_wang,ding2022transmvsnet,wang2022itermvs,wang2022efficient} use a constant search range for all pixels to generate new depth hypotheses around the previous estimation. 
However, it is reasonable to include confidence/uncertainty of the estimation in the sampling. Specifically, the search range should be reduced for pixels with accurate estimation to further improve accuracy, while enlarged for pixels with erroneous estimation so that the search range may cover the ground truth. 
For example, UCSNet~\cite{cheng_2020_ucsnet} computes the per-pixel sampling range with the variance of the probability volume on multiple stages. 
However, this is not suitable in our framework since we do not estimate the probability volume during refinement. 

In our methods, we adaptively adjust the per-pixel sampling range $\bm{R}_{k}$ based on the confidence $\bm{C}_{k-1}$ from the diffusion model. For the first iteration $k=1$, we set $\bm{R}_{1} = \bm{R}_{\text{init}}$, where $\bm{R}_{\text{init}}$ is a pre-defined range for each stage. For $k>1$, $\bm{R}_{k}$ is computed as: 
\begin{equation}\label{eq:conf_range}
    \bm{R}_{k} = (1 - \bm{C}_{k-1}) \cdot (\bm{R}_{\text{max}} - \bm{R}_{\text{min}}) + \bm{R}_{\text{min}},
\end{equation}
where $\bm{R}_{\text{min}}=\lambda_{\text{min}} \cdot \bm{R}_{\text{init}}$, $\bm{R}_{\text{max}}=\lambda_{\text{max}} \cdot \bm{R}_{\text{init}}$ are pre-defined limits of the sampling range.  
Then we uniformly sample $D_1$ depth hypotheses in the inverse range $[\bm{\bar{D}}_{k-1}-\bm{R}_{k}, \bm{\bar{D}}_{k-1}+\bm{R}_{k}]$.

\subsection{Learned upsampling}
Since our methods use multi-stage structure, we need to upsample the depth between different stages. 
Instead of simple bilinear or nearest interpolation~\cite{gu_2020_cascademvsnet,patchmatchnet_wang}, we use learned upsampling with mask~\cite{teed2020raft,wang2022itermvs}, which experimentally improves performance. Specifically, we compute the depth value of each pixel in the upsampled depth as the convex combination of a $3\times3$ grid of its coarse resolution neighbors. 
For stage $m$, we predict a mask, with a shape as $H_m \times W_m \times (r \times r \times 9)$, by applying 2D convolution layers on the reference context feature $\bm{F}_c^m$, where $r$ is the upsample ratio. Then we apply \textit{softmax} over the weights of the 9 neighbors and predict the upsampled depth as the weighted combination over the neighborhood.

\subsection{Training loss}
Our loss function $\mathcal{L}_{\text{full}}$ includes the losses for all the depth maps,~\ie, initial depth $\bm{D}_{\text{init}}$, intermediate depth inside the diffusion-based refinement $\{\bm{D}_{t,k}\}_{k=1}^K$ at timestep $t$, and all the upsampled depth maps. 
For clarification, we sort all depth maps based on the estimation order as $\{\bm{D}_j\}_{j=1}^J$, where $J$ is the number of depth maps. 
Following IterMVS~\cite{wang2022itermvs}, we transform the depth map $\bm{D}$ into the normalized inverse space with Eq.~\ref{eq::normalization} and compute the $L_1$ loss w.r.t. the ground truth depth map $\bm{D}_{\text{gt}}$: 
\begin{equation}
    \mathcal{L} = | \bm{\bar{D}} - \bm{\bar{D}}_{\text{gt}} |,
\end{equation}
where $\bm{\bar{D}}, \bm{\bar{D}}_{\text{gt}}$ are the normalized inverse depth for $\bm{D},\bm{D}_{\text{gt}}$. 
For the depth map $\bm{D}$ in diffusion-based refinement where we estimate the corresponding confidence $\bm{C}$, we include $\bm{C}$ inside the loss function as: 
\begin{equation}\label{eq:loss_conf}
    \mathcal{L} = \frac{| \bm{\bar{D}} - \bm{\bar{D}}_{\text{gt}} |}{1 - \bm{C}} + \lambda_C \cdot \log(1 - \bm{C}),
\end{equation}
where $\lambda_C=0.05$ is the weight. 
We use confidence $\bm{C}$ to adjust the weight for $L_1$ loss, and $ \lambda_C \cdot \log(1 - \bm{C})$ as a regularization term to avoid trivial solution as $\bm{C}=0$. 

Following~\cite{teed2020raft,wang2022itermvs,wang2022efficient}, we use exponentially increasing weights for the depth maps to balance the depth supervisions across different stages and iterations. The motivation is that the depth should be estimated from coarse to fine and thus the error is gradually penalized more. Therefore, the final loss $\mathcal{L}_{\text{full}}$ is written as:
\begin{equation}
    \mathcal{L}_{\text{full}} = \sum_{j=1}^{J} \beta^{J-j} \mathcal{L}_j,
\end{equation}
where $\mathcal{L}_j$ is the loss for $\bm{D}_j$, $\beta=0.9$ is the weight. %

\section{Experiments}

\subsection{Datasets}
Following prior works, we use DTU~\cite{aanaes2016_dtu} and BlendedMVS~\cite{yao2020blendedmvs} for training, and evaluate the performance on DTU, Tanks \& Temples~\cite{knapitsch2017tanks} and ETH3D~\cite{2017eth3d}. 

\customparagraph{DTU dataset}
DTU~\cite{aanaes2016_dtu} is an indoor multi-view stereo dataset collected with known accurate camera trajectory. It contains 124 different object-centered scenes captured with 7 different lighting conditions. 
Following previous methods, we use the training, testing and validation sets introduced in SurfaceNet~\cite{ji_2017_surfacenet}. 

\customparagraph{BlendedMVS}
BlendedMVS~\cite{yao2020blendedmvs} is a large-scale synthetic dataset with various scenes. It provides over 17$k$ high-quality training samples with accurate depth maps and camera poses. It is divided into the training set and validation set. Compared with DTU~\cite{aanaes2016_dtu}, BlendedMVS contains scenes with various scales,~\eg, from object level to city level, and is thus commonly used to improve the generalization ability of learning-based MVS methods. 

\customparagraph{Tanks \& Temples}
Tanks \& Temples~\cite{knapitsch2017tanks} is a large-scale real-world dataset consisting of both indoor and outdoor scenes, which is divided into training set
and test set. Learning-based MVS methods~\cite{yao_2018_mvsnet,gu_2020_cascademvsnet,patchmatchnet_wang} commonly test the zero-shot generalization ability on the test set. The test set is further divided into \textit{intermediate} and \textit{advanced} subsets, where the \textit{advanced} subset is more challenging than the \textit{intermediate} subset because of the complex structures. 

\customparagraph{ETH3D}
ETH3D~\cite{2017eth3d} is a large-scale dataset captured in complex real-world scenes under challenging conditions,~\eg, low-textured regions and wide baselines. It contains both outdoor and indoor scenes. The dataset is divided into training set and test set. Both sets are commonly used to test the zero-shot generalization ability of learning-based MVS methods~\cite{ma2021epp,wang2022itermvs,wang2022mvster}.

\subsection{Implementations}
In this section, we discuss the implementation details of DiffMVS and CasDiffMVS, including hyper-parameter settings, training and inference details. 

\customparagraph{Hyper-parameters}
For depth initialization, we set the number of initial samples $D_{0}=48$ for both DiffMVS and CasDiffMVS. 
For confidence-based sampling, we set $D_{1}=6$, $\lambda_{\text{min}}=0.25$, $\lambda_{\text{max}}=4$, $\bm{R}_{\text{init}} =3 / 192.0$ for DiffMVS. For CasDiffMVS, we set $\bm{R}_{\text{init}}=1 / 96.0$ for stage 2 and $\bm{R}_{\text{init}}=1 / 192.0$ for stage 3. We set $D_{1}=4$, $\lambda_{\text{min}}=0.125$, $\lambda_{\text{max}}=8$ for both stages of CasDiffMVS.

For diffusion models, we set the total diffusion timesteps as $T=1000$. In each diffusion timestep, we refine $K=4$ times for DiffMVS and $K=3$ times for CasDiffMVS. 
Since we aim to refine the coarse initial depth and \textit{normalize} the depth values during diffusion into the range $[-1, 1]$, we find that using standard Gaussian noise,~\ie, $\mathcal{N} (0, \bm{I})$, like DDPM~\cite{ho2020denoising} will introduce too strong noise for refinement. 
Therefore, we set the Gaussian noise for stage $m$ as $\mathcal{N} (0, \sigma_m^2 \bm{I})$ to control the noise scale. Specifically, we set $\sigma_2=0.5$ for the refinement on stage 2, and $\sigma_3=0.1$ for the refinement on stage 3 when training on DTU. 
Considering that the DTU-trained models provide a good initialization for fine-tuning and the domain gap between the DTU and large-scale BlendedMVS, we further design a \textit{noise-scaling strategy} to fine-tune our models on BlendedMVS. Specifically, $\sigma_2$ and $\sigma_3$ will be halved at the start of fine-tuning and after 8 epochs. We empirically find this enables our models to generalize better on Tanks \& Temples~\cite{knapitsch2017tanks} and ETH3D~\cite{2017eth3d}.

\customparagraph{Training} 
We implement DiffMVS and CasDiffMVS with PyTorch~\cite{paszke2019pytorchai}, and use Adam~\cite{kingma2015adam} ($\beta_1 \!=\! 0.9, \beta_2 \!=\! 0.999$) as the optimizer. 
First, we train our models on DTU training set~\cite{aanaes2016_dtu} for evaluation on DTU testing set. 
For evaluation on Tanks \& Temples~\cite{knapitsch2017tanks} and ETH3D~\cite{2017eth3d}, 
we further finetune the DTU-pretrained models on BlendedMVS~\cite{yao2020blendedmvs}. 
The image resolution is set to $640 \!\times\! 512$ for DTU, and $768 \!\times\! 576$ for BlendedMVS. The number of input views $N$ is set to 5 for DTU and 9 for BlendedMVS. 
On both datasets, we set the batch size as 4 and train the models under OneCycleLR scheduler with a maximum learning rate as 0.001. We train DiffMVS and CasDiffMVS for 12 and 16 epochs respectively on each dataset.

\customparagraph{Inference} 
During inference, we adopt DDIM~\cite{song2020denoising} to improve sampling efficiency. We set DDIM sampling timestep as $T_s = 1$ since we observe that our methods converge fast with 1 sampling timestep only and the performance will not explicitly improve with more sampling timesteps.

After depth estimation, we use photometric and geometric consistency to filter outliers in the depth maps, following common practices~\cite{yao_2018_mvsnet,d2hcrmvsnet}, and back-project the valid pixels into the 3D space as the final point cloud. %

\subsection{Evaluation}
In this section, we evaluate both DiffMVS and CasDiffMVS on popular MVS benchmarks~\cite{aanaes2016_dtu,knapitsch2017tanks,2017eth3d} and compare with state-of-the-art methods. In addition, since efficiency in run-time and memory consumption is important in practice, we compare the efficiency of our methods with state-of-the-art methods. 

\customparagraph{Evaluation on DTU}
Following prior works, we evaluate with our models trained on DTU training set only. 
We set the image size and number of views $N$ to $1600 \!\times\! 1152$ and 5 respectively. 
Table~\ref{tab:evaluation_dtu} summarizes the quantitative results. 
Compared with IterMVS~\cite{wang2022itermvs}, the current state-of-the-art method with single-stage refinement, DiffMVS outperforms it with a large margin in all metrics. Moreover, DiffMVS performs better than many learning-based methods with multi-stage refinement,~\eg, Effi-MVS~\cite{wang2022efficient}, UniMVSNet\cite{peng2022rethinking}, in \textit{overall} quality. %
As the multi-stage extension of DiffMVS, CasDiffMVS outperforms other learning-based methods in \textit{accuracy} and achieves very competitive performance in \textit{overall} quality when compared with the top-performing methods~\cite{zhang2023geomvsnet,liu2023epipolar}. 

\begin{table}[t]
\centering
\setlength{\tabcolsep}{2.5pt}
\footnotesize

\caption{Quantitative results on DTU~\cite{aanaes2016_dtu}. 
Methods are separated into four categories (from top to bottom): traditional methods, learning-based methods without refinement, with single-stage refinement and with multi-stage refinement. 
Red, orange, and yellow highlights indicate the 1st, 2nd, and 3rd-best performing method. 
}
\begin{tabular}{lccc}
 \hline
 Methods & Acc.(mm) $\downarrow$ & Comp.(mm) $\downarrow$ & Overall(mm) $\downarrow$\\
 \hline
 Gipuma~\cite{galliani_2015_gipuma} & \cellcolor{cred}0.283 & 0.873 & 0.578 \\
 COLMAP~\cite{schonberger2016pixelwise} & 0.400 & 0.664 & 0.532 \\
 \hline
 MVSNet~\cite{yao_2018_mvsnet} & 0.396 & 0.527 & 0.462\\
 R-MVSNet~\cite{yao_2019_rmvsnet} & 0.383 & 0.452 & 0.417\\
 AA-RMVSNet~\cite{aarmvsnet} & 0.376 & 0.339 & 0.357 \\
 \hline
 IterMVS~\cite{wang2022itermvs} & 0.373 & 0.354 & 0.363 \\
  \textbf{DiffMVS} & \cellcolor{yellow}0.318 & 0.297 & 0.308 \\
 \hline
 CasMVSNet~\cite{gu_2020_cascademvsnet} & 0.325 & 0.385 & 0.355\\
 PatchmatchNet~\cite{patchmatchnet_wang} & 0.427 & 0.277 & 0.352 \\
 EPP-MVSNet~\cite{ma2021epp} & 0.413 & 0.296 & 0.355 \\
 PVSNet~\cite{xu2022learning} & 0.337 & 0.315 & 0.326 \\
 UniMVSNet~\cite{peng2022rethinking} & 0.352 & 0.278 & 0.315 \\
 TransMVSNet~\cite{ding2022transmvsnet} & 0.321 & 0.289 & 0.305 \\
 Effi-MVS~\cite{wang2022efficient} & 0.321 & 0.313 & 0.317 \\
 MVSTER~\cite{wang2022mvster} & 0.340 & {0.266} & 0.303 \\
 GeoMVSNet~\cite{zhang2023geomvsnet} & 0.331 & \cellcolor{yellow}0.259 & \cellcolor{orange}0.295 \\
 ET-MVSNet~\cite{liu2023epipolar} & 0.329 & \cellcolor{orange}0.253 & \cellcolor{cred}0.291 \\
 EI-MVSNet~\cite{chang2024ei} & 0.346 & 0.260 & 0.303 \\
 Effi-MVS+~\cite{wang2024efficient} & 0.327 & 0.275 & 0.301 \\
 GC-MVSNet~\cite{vats2024gc} & 0.330 & 0.260 & \cellcolor{orange}0.295 \\
 CANet~\cite{su2025context} & 0.351 & \cellcolor{cred}0.248 & 0.299 \\
  \textbf{CasDiffMVS} & \cellcolor{orange}0.310 & 0.286 & \cellcolor{yellow}{0.298} \\
 \hline
\end{tabular}
  \setlength{\belowcaptionskip}{-10pt}
\label{tab:evaluation_dtu}
\end{table}

\begin{figure*}[tbp]
\centering
{\includegraphics[width=1.0\linewidth]{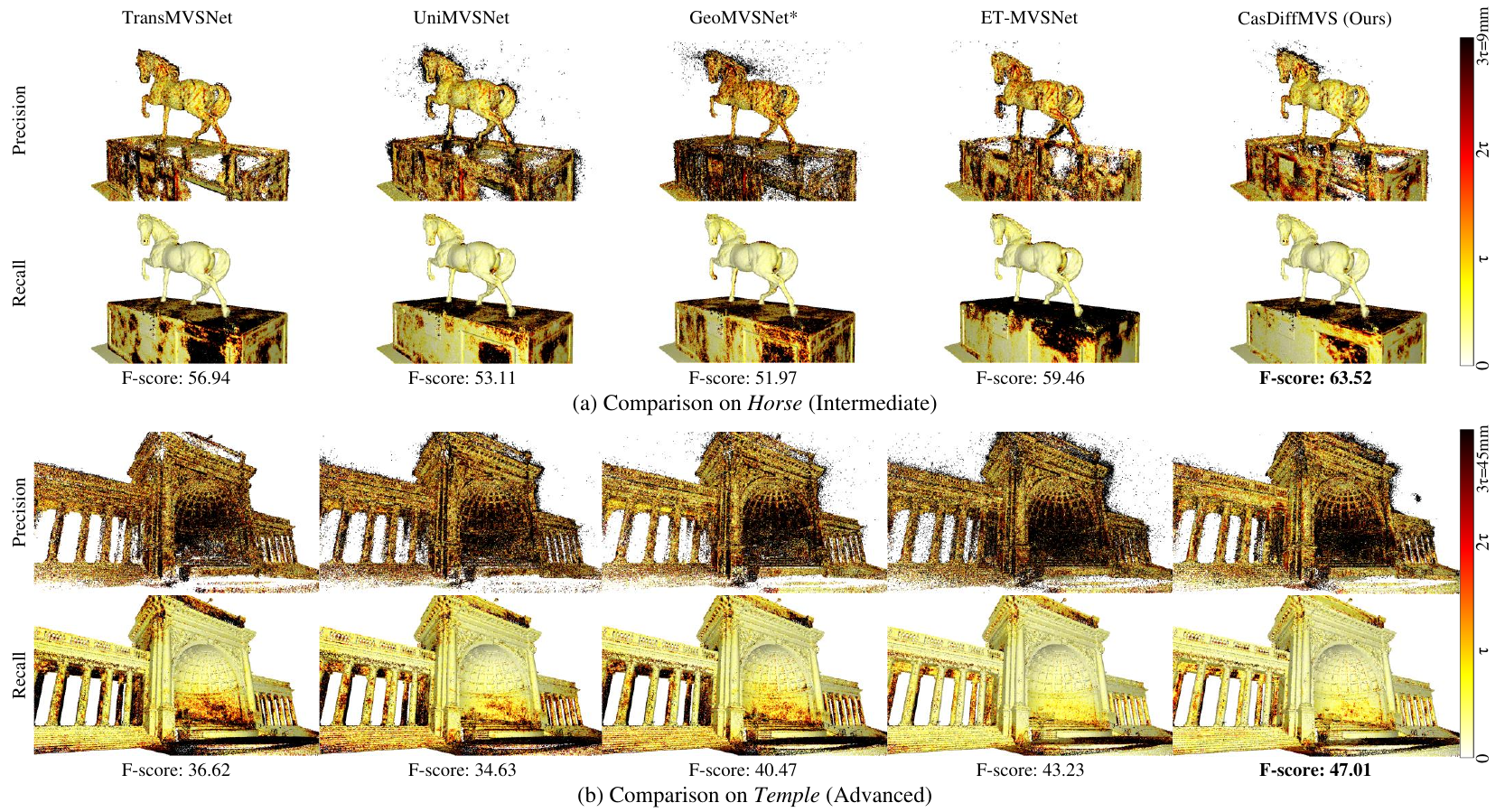}}
\caption{
Qualitative comparisons of reconstruction errors on Tanks and Temples~\cite{knapitsch2017tanks}. We visualize precision and recall error maps for `Horse' and `Temple' scenes. 
}\label{fig:tanks_comp}
\end{figure*}

\customparagraph{Evaluation on Tanks \& Temples}
To demonstrate the zero-shot generalization capability of our methods, we evaluate on Tanks \& Temples. 
We use the depth range, camera parameters and view selection provided by PatchmatchNet~\cite{patchmatchnet_wang}. 
The image size and number of views are set to $1920 \!\times\! 1056$ and 10 respectively.  
For depth filtering, we use dynamic geometric consistency checking method introduced in~\cite{d2hcrmvsnet}. 
The quantitative results are summarized in Table~\ref{tab:evaluation_tank}. 
For the learning-based methods with single-stage refinement, our DiffMVS is 11.33\% and 16.15\% better than IterMVS~\cite{wang2022itermvs} on intermediate and advanced set respectively.  
Remarkably, it also outperforms some learning-based methods with multi-stage refinement, such as PVSNet~\cite{xu2022learning} and MVSTER~\cite{wang2022mvster}. 
For the learning-based methods with multi-stage refinement, our CasDiffMVS achieves state-of-the-art performance on both sets. 
In Fig.~\ref{fig:tanks_comp}, we visualize the reconstruction errors on `Horse' and `Temples' scenes. CasDiffMVS demonstrates significant improvements over existing methods and produces more complete surfaces.  
Overall, our methods demonstrate very competitive generalization performance.

\begin{table*}
\centering
\caption{Quantitative results on Tanks \& Temples~\cite{knapitsch2017tanks} using F-score (\%) (higher is better). Methods are separated into four categories (from top to bottom): traditional methods, learning-based methods without refinement, with single-stage refinement and with multi-stage refinement. Red, orange, and yellow highlights indicate the 1st, 2nd, and 3rd-best performing method. 
}
\resizebox{\linewidth}{!}{
\begin{tabular}{l|c|cccccccc|c|cccccc}
 \hline
 \multirow{2}{*}{Methods} & \multicolumn{9}{c|}{Intermediate}  & \multicolumn{7}{c}{Advanced} \\
 \cline{2-17}
& Mean & Family & Francis & Horse & Light. & M60 & Panther & Play. & Train & Mean & Audi. & Ball. & Court. & Museum & Palace & Temple \\
 \hline
 COLMAP~\cite{schoenberger2016colmap} & 42.14 & 50.41 & 22.25 & 26.63 & 56.43 & 44.83 & 46.97 & 48.53 & 42.04 & 27.24 & 16.02 & 25.23 & 34.70 & 41.51 & 18.05 & 27.94 \\
 ACMM~\cite{xu_2019_acmm} & 57.27 & 69.24 & 51.45 & 46.97 & 63.20 & 55.07 & 57.64 & 60.08 & 54.48 & 34.02 & 23.41 & 32.91 & \cellcolor{cred}41.17 & 48.13 & 23.87 & 34.60 \\
 HPM-MVS~\cite{ren2023hierarchical} & 61.39	& 73.40	& 57.67 & 56.96	& 64.70	& 60.39	& 61.21	& 60.35 & 56.43 & 40.80 & \cellcolor{cred}32.85 & 46.00 & \cellcolor{orange}40.92 & \cellcolor{cred}53.04 & 29.63 & 42.37 \\
 \hline
 R-MVSNet~\cite{yao_2019_rmvsnet} & 48.40 & 69.96 & 46.65 & 32.59 & 42.95 & 51.88 & 48.80 & 52.00 & 42.38 & 24.91 & 12.55 & 29.09 & 25.06 & 38.68 & 19.14 & 24.96 \\
 \hline
 IterMVS~\cite{wang2022itermvs} & 56.94 & 76.12 & 55.80 & 50.53 & 56.05 & 57.68 & 52.62 & 55.70 & 50.99 & 34.17 & 25.90 & 38.41 & 31.16 & 44.83 & 29.59 & 35.15 \\
 \textbf{DiffMVS} & 63.39 & 78.14 & 62.73 & \cellcolor{orange}62.29 & 63.20 & 61.10 & 61.84 & 59.65 & 58.18 & 39.69 & 31.10 & 43.45 & 37.85 & 48.74 & 32.94 & \cellcolor{yellow}44.05 \\
 \hline
 CasMVSNet~\cite{gu_2020_cascademvsnet} & 56.84 & 76.37 & 58.45 & 46.26 & 55.81 & 56.11 & 54.06 & 58.18 & 49.51 & 31.12 & 19.81 & 38.46 & 29.10 & 43.87 & 27.36 & 28.11 \\
 PatchmatchNet~\cite{patchmatchnet_wang} & 53.15 & 66.99 & 52.64 & 43.24 & 54.87 & 52.87 & 49.54 & 54.21 & 50.81 & 32.31 & 23.69 & 37.73 & 30.04 & 41.80 & 28.31 & 32.29 \\
 EPP-MVSNet~\cite{ma2021epp} & 61.68 & 77.68 & 60.54 & 52.96 & 62.33 & 61.69 & 60.34 & \cellcolor{cred}62.44 & 55.30 & 35.72 & 21.28 & 39.74 & 35.34 & 49.21 & 30.00 & 38.75 \\
 PVSNet~\cite{xu2022learning} & 59.11 & 78.13 & 61.62 & 52.11 & 56.90 & 60.12 & 53.77 & 57.58 & 52.64 & 35.51 & 24.40 & 40.96 & 34.23 & 47.95 & 29.02 & 36.50 \\
 UniMVSNet~\cite{peng2022rethinking} & 64.36 & 81.20 & 66.43 & 53.11 & 63.46 & \cellcolor{cred}66.09 & \cellcolor{cred}64.84 & \cellcolor{orange}62.23 & 57.53 & 38.96 & 28.33 & 44.36 & 39.74 & \cellcolor{orange}52.89 & 33.80 & 34.63 \\
 TransMVSNet~\cite{ding2022transmvsnet} & 63.52 & 80.92 & 65.83 & 56.94 & 62.54 & 63.06 & 60.00 & 60.20 & 58.67 & 37.00 & 24.84 & 44.59 & 34.77 & 46.49 & 34.69 & 36.62 \\
 Effi-MVS~\cite{wang2022efficient} & 56.88 & 72.21 & 51.05 & 51.78 & 58.63 & 58.71 & 56.21 & 57.07 & 49.38 & 34.39 & 20.22 & 42.39 & 33.73 & 45.08 & 29.81 & 35.09 \\
 MVSTER~\cite{wang2022mvster} & 60.92 & 80.21 & 63.51 & 52.30 & 61.38 & 61.47 & 58.16 & 58.98 & 51.38 & 37.53 & 26.68 & 42.14 & 35.65 & 49.37 & 32.16 & 39.19 \\
 GeoMVSNet*~\cite{zhang2023geomvsnet} & 62.67 & 80.12 & 66.14 & 51.97 & \cellcolor{yellow}65.95 & 61.70 & 60.40 & 61.20 & 53.91 & 39.08 & 25.30 & 45.75 & 37.72 & 50.36 & 34.85 & 40.47 \\
 ET-MVSNet~\cite{liu2023epipolar} & \cellcolor{yellow}65.49 & \cellcolor{orange}81.65 & \cellcolor{orange}68.79 & 59.46 & 65.72 & 64.22 & \cellcolor{yellow}64.03 & 61.23 & \cellcolor{yellow}58.79 & 40.41 & 28.86 & 45.18 & 38.66 & 51.10 & \cellcolor{yellow}35.39 & 43.23 \\
 EI-MVSNet~\cite{chang2024ei} & \cellcolor{orange}65.52 & \cellcolor{yellow}81.59 & \cellcolor{yellow}67.67 & \cellcolor{yellow}61.67 & 63.18 & \cellcolor{orange}65.10 & 63.42 & 60.62 & \cellcolor{cred}60.95 & 40.68 & 29.97 & 45.86 & 38.45 & 49.50 & \cellcolor{orange}35.78 & \cellcolor{orange}44.53 \\
 Effi-MVS+~\cite{wang2024efficient} & 64.07 & 79.87 & 66.77 & 57.29 & \cellcolor{orange}66.35 & 62.83 & 61.11 & \cellcolor{yellow}62.14 & 56.23 & \cellcolor{yellow}41.20 & \cellcolor{yellow}32.04 & \cellcolor{cred}47.04 & 38.84 & 51.26 & 34.95 & 43.06 \\
 GC-MVSNet~\cite{vats2024gc} & 62.74 & 80.87 & 67.13 & 53.82 & 61.05 & 62.60 & 59.64 & 58.68 & 58.48 & 38.74 & 25.37 & \cellcolor{orange}46.50 & 36.65 & 49.97 & \cellcolor{cred}35.81 & 38.11 \\
 CANet~\cite{su2025context} & 65.05 & 80.41	& 63.85	& 59.62 & \cellcolor{cred}67.32 & \cellcolor{yellow}65.03 & \cellcolor{orange}64.18 & 62.05 & 57.90 & \cellcolor{orange}41.22 & 31.26 & \cellcolor{yellow}46.17 & \cellcolor{yellow}40.54 & \cellcolor{yellow}52.70 & 33.49	& 43.15 \\
 \textbf{CasDiffMVS} & \cellcolor{cred}65.87 & \cellcolor{cred}81.74 & \cellcolor{cred}69.21 & \cellcolor{cred}63.52 & 65.89 & 62.92 & 62.35 & 61.31 & \cellcolor{orange}60.00 & \cellcolor{cred}41.81 & \cellcolor{orange}32.66 & 45.70 & 39.34 & 50.93 & 35.25 & \cellcolor{cred}47.01 \\
 \hline

\end{tabular}}

\footnotesize{*: GeoMVSNet does not provide official checkpoint on BlendedMVS. Following official scripts, we finetune the DTU pre-trained model on BlendedMVS. }
\label{tab:evaluation_tank}
\end{table*}

\begin{figure*}[tbp]
\centering
{\includegraphics[width=1.0\linewidth]{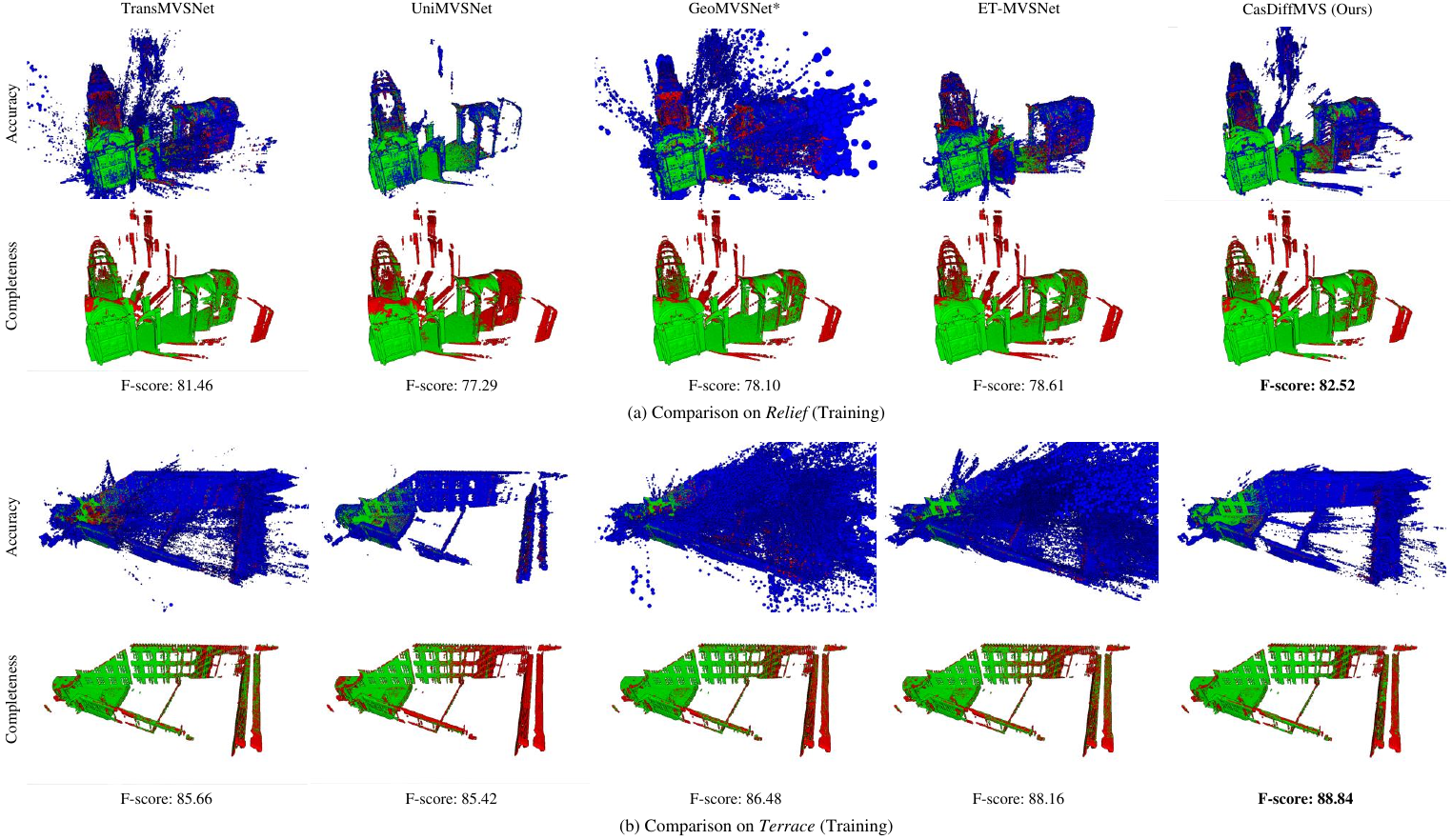}}
\caption{
Qualitative comparisons of reconstruction errors on ETH3D~\cite{2017eth3d}. We visualize accuracy and completeness error maps for indoor `Relief' and outdoor `Terrace' scenes. Green points are accurate, red points are inaccurate and blue points are unobserved with respect to the ground truth. 
}\label{fig:eth3d_comp}
\end{figure*}

\customparagraph{Evaluation on ETH3D}
We further evaluate the generalization ability of our methods on the challenging ETH3D. 
We set the image size and number of views to $1920 \!\times\! 1280$ and 10 respectively. 
The results are summarized in Table~\ref{tab:evaluation_eth}. 
Since ETH3D is a large-scale dataset with large baselines, it poses great challenges to the generalization ability of learning-based methods. Traditional PatchMatch MVS methods~\cite{xu2022multi,ren2023hierarchical} leverage the fast depth search and random depth perturbations to achieve promising results on this dataset. By introducing the diffusion denoising process, our methods can better consider depth sampling on this dataset. 
Therefore, our CasDiffMVS achieves the best performance among the learning-based methods on both training and test sets. Moreover, compared with HPM-MVS~\cite{ren2023hierarchical}, the state-of-the-art traditional method, CasDiffMVS achieves competitive performance on the test set.   
In addition, compared with IterMVS~\cite{wang2022itermvs}, our DiffMVS outperforms it on both training set and test set. Note that, DiffMVS also performs better than many state-of-the-art multi-stage methods, \eg, GeoMVSNet~\cite{zhang2023geomvsnet}, ET-MVSNet~\cite{liu2023epipolar}, Effi-MVS+~\cite{wang2024efficient}, on both sets. 
Fig.~\ref{fig:eth3d_comp} shows the reconstruction error comparisons for `Relief' and `Terrace' scenes.  CasDiffMVS achieves more accurate and complete reconstructions than other methods. 
These results further demonstrate the generalization capabilities of our methods in challenging scenarios.

\begin{table*}[t]
\centering
\caption{Quantitative results of different methods on ETH3D~\cite{2017eth3d} using $F_1$-score (at evaluation threshold $2cm$, higher is better).  Methods are separated into three categories (from top to bottom): traditional methods, learning-based methods with single-stage refinement and with multi-stage refinement. 
}
\begin{tabular}{l|ccc|ccc}
 \hline
 \multirow{2}{*}{Methods} & \multicolumn{3}{c|}{Training}  & \multicolumn{3}{c}{Test} \\
  \cline{2-7}
  & Acc. $\uparrow$ & Comp. $\uparrow$ & $F_1$-score $\uparrow$ & Acc. $\uparrow$ & Comp. $\uparrow$ & $F_1$-score $\uparrow$ \\
 \hline
 COLMAP~\cite{schoenberger2016colmap} & \cellcolor{cred}91.85 & 55.13 & 67.66 & \cellcolor{orange}91.97 & 62.98 & 73.01 \\
 ACMM~\cite{xu_2019_acmm} & \cellcolor{orange}90.67 & 70.42 & \cellcolor{orange}78.86 & 90.65 & 74.34 & 80.78 \\
 HPM-MVS~\cite{ren2023hierarchical} & \cellcolor{yellow}90.66 & \cellcolor{cred}79.50 & \cellcolor{cred}84.58 &  \cellcolor{cred}92.13 & 83.25 & \cellcolor{cred}87.11 \\
 \hline
 IterMVS~\cite{wang2022itermvs} & 79.79 & 66.08 & 71.69 & 84.73 & 76.49 & 80.06 \\
 \textbf{DiffMVS} & 76.74 & 74.32 & 74.86 & 80.40 & \cellcolor{orange}84.28 & 82.10 \\
 \hline
 PatchmatchNet~\cite{patchmatchnet_wang} & 64.81 & 65.43 & 64.21 & 69.71 & 77.46 & 73.12 \\
 EPP-MVSNet~\cite{ma2021epp} & 82.76 & 67.58 & 74.00 & 85.47 & 81.49 & 83.40 \\
 PVSNet~\cite{xu2022learning} & 83.00 & 71.76 & 76.57 & 81.55 & 83.97 & 82.62 \\
 UniMVSNet~\cite{peng2022rethinking} & 85.39 & 57.83 & 67.18 & \cellcolor{yellow}89.12 & 72.74 & 79.10 \\
 TransMVSNet~\cite{ding2022transmvsnet} & 69.62 & 71.47 & 70.10 & 73.26 & 81.84 & 76.98 \\
 MVSTER~\cite{wang2022mvster} & 68.08 & \cellcolor{orange}76.92 & 72.06 & 77.09 & 82.47 & 79.01 \\
 GeoMVSNet*~\cite{zhang2023geomvsnet} & 68.71 & 71.22 & 69.69 & 69.77 & 83.68 & 75.70 \\
 ET-MVSNet~\cite{liu2023epipolar} & 74.71 & 71.38 & 72.46 & 75.28 & \cellcolor{yellow}84.01 & 78.63 \\
 EI-MVSNet~\cite{chang2024ei} & - & - & - & 85.12 & 83.77 & \cellcolor{yellow}84.19 \\
 Effi-MVS+~\cite{wang2024efficient} & 82.33 & 70.42 & 75.28 & 83.90 & 83.81 & 83.62 \\
 \textbf{CasDiffMVS} & 79.93 & \cellcolor{yellow}75.20 & \cellcolor{yellow}76.76 & 85.21 & \cellcolor{cred}85.37 & \cellcolor{orange}85.11 \\
 \hline

\end{tabular}

\label{tab:evaluation_eth}
\end{table*}

\begin{figure}[tbp]
\centering
{\includegraphics[width=1.0\columnwidth]{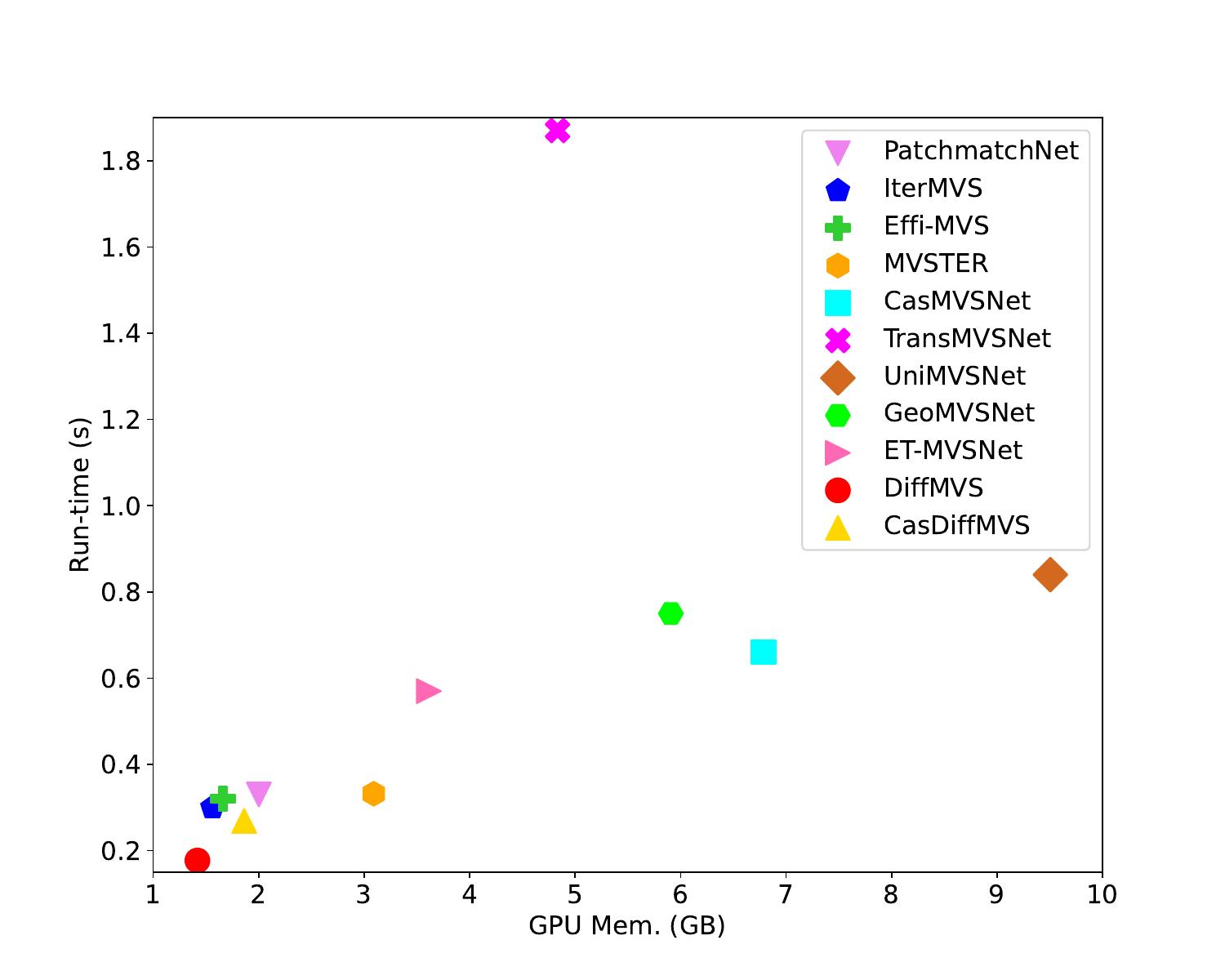}}
\caption{
Efficiency comparison with state-of-the-art MVS methods~\cite{gu_2020_cascademvsnet, patchmatchnet_wang,wang2022itermvs,wang2022efficient,wang2022mvster,ding2022transmvsnet,peng2022rethinking,zhang2023geomvsnet,liu2023epipolar} on DTU~\cite{aanaes2016_dtu} (image size: $1600 \times 1152$, number of input views $N=5$). For fair comparison, all experiments are done on one workstation with a NVIDIA 2080 Ti GPU. 
}\label{fig:efficiency}
\end{figure}

\customparagraph{Efficiency comparison}
Efficiency in memory and run-time is important in industrial applications, especially for mobile devices with limited computational resources. 
Therefore, we compare the run-time and GPU memory consumption of our methods with the state-of-the-art MVS methods on a workstation with one NVIDIA 2080 Ti GPU and visualize the results in Fig.~\ref{fig:efficiency}. 
Our DiffMVS achieves highest efficiency in both run-time and GPU memory. 
Compared with IterMVS~\cite{wang2022itermvs}, the current most efficient method, DiffMVS consumes 9.13\% less GPU memory and is 69.49\% faster. 
Moreover, on DTU~\cite{aanaes2016_dtu}, Tanks \& Temples~\cite{knapitsch2017tanks} and ETH3D~\cite{2017eth3d}, DiffMVS mostly outperforms state-of-the-art efficient methods~\cite{patchmatchnet_wang,wang2022itermvs,wang2022efficient,wang2022mvster} and achieves competitive performance as TransMVSNet~\cite{ding2022transmvsnet} and UniMVSNet~\cite{peng2022rethinking}, while being much more efficient. 
CasDiffMVS has more computational overheads than DiffMVS because of the two-stage diffusion-based refinement. However, CasDiffMVS still achieves similar efficiency as PatchmatchNet~\cite{patchmatchnet_wang} in both GPU memory and run-time, while being more efficient than the top-performing methods~\cite{ding2022transmvsnet,peng2022rethinking,zhang2023geomvsnet,liu2023epipolar}. Moreover, CasDiffMVS achieves very competitive performance on three benchmarks when compared with these top-performing methods~\cite{ding2022transmvsnet,peng2022rethinking,zhang2023geomvsnet,liu2023epipolar}.

\begin{figure}[tbp]
\centering
{\includegraphics[width=1.0\columnwidth]{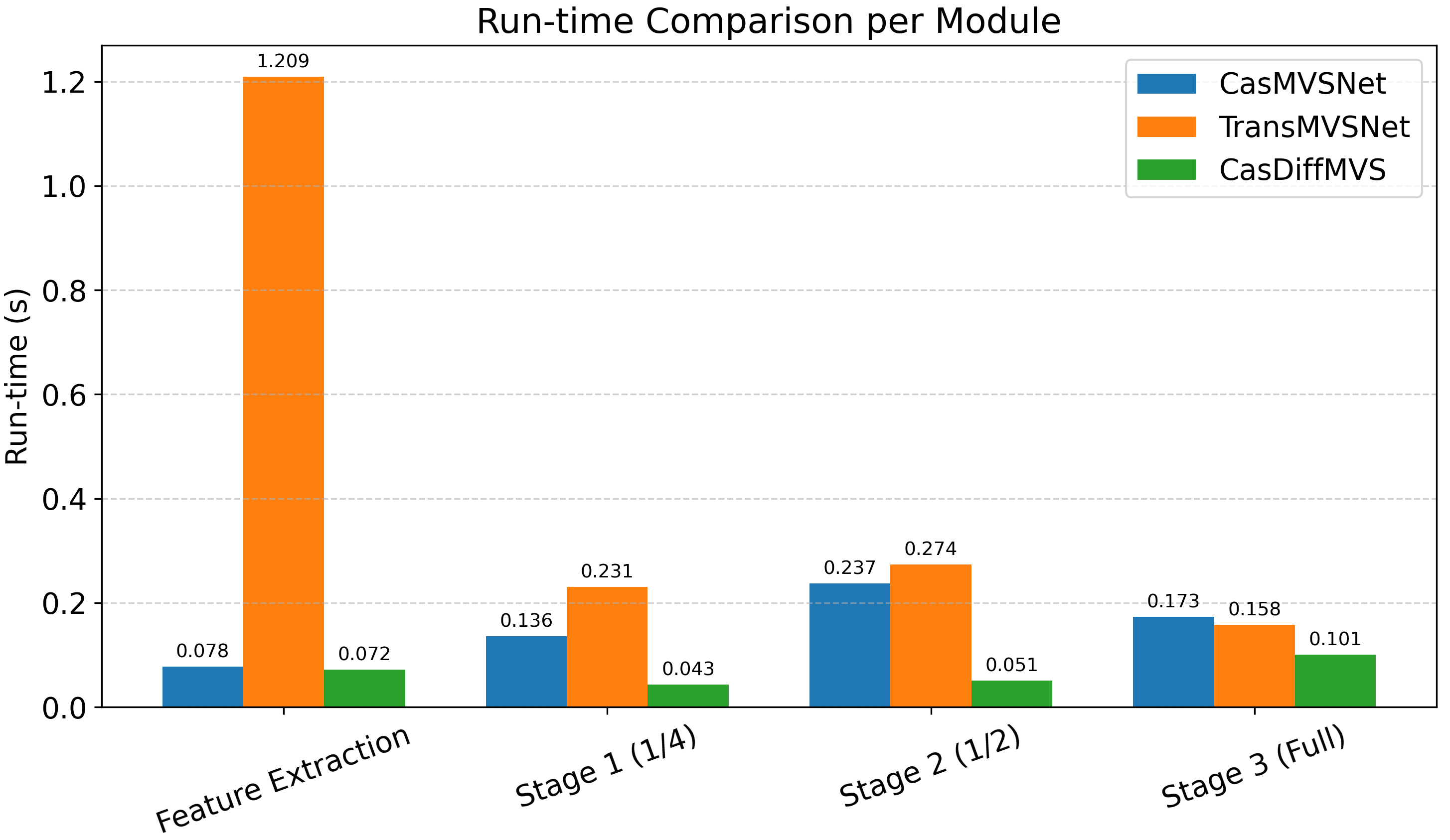}}
\caption{
Run-time comparison per module on DTU~\cite{aanaes2016_dtu} (image size: $1600 \times 1152$, number of input views $N=5$). For fair comparison, all experiments are done on one workstation with a NVIDIA 2080 Ti GPU. 
}\label{fig:runtime_bar}
\end{figure}

To better illustrate how our proposed method CasDiffMVS achieves its runtime advantages over 3D CNN or transformer-based frameworks, we show run-time comparison per module for CasMVSNet, TransMVSNet and CasDiffMVS in Fig.~\ref{fig:runtime_bar}. We observe that TransMVSNet is the slowest because of the expensive attention operation introduced in feature extraction. For depth inference in different stages, CasDiffMVS runs much faster than CasMVSNet and TransMVSNet, where both CasMVSNet and TransMVSNet use 3D CNN. 
Compared to these methods with 3D CNN and attention modules that are computationally expensive, we carefully design the network architecture to reduce computation,~\textit{e.g.} no 3D CNN or attention. In addition, since our framework predicts an initial depth map $\bm{D}_{\text{init}}$ and then uses a diffusion model to refine it, our framework requires fewer sampling timesteps (See more analysis in the next section). As a result, our CasDiffMVS is more efficient than 3D CNN or transformer-based frameworks.

\subsection{Ablation study}
In this section, we conduct an ablation study to validate the effectiveness of different components in our pipeline. If not specified, the experiments are conducted with DiffMVS. For evaluation on DTU~\cite{aanaes2016_dtu} testing set, we use the models trained on DTU training set only. For evaluation on ETH3D~\cite{2017eth3d} training set, we use the models finetuned on BlendedMVS~\cite{yao2020blendedmvs}. %

\begin{table*}[t]
\centering
\caption{Ablation study of DiffMVS on DTU~\cite{aanaes2016_dtu} and ETH3D~\cite{2017eth3d}. Settings used in our method are underlined. 
}
\begin{tabular}{c|c|ccc|ccc}
 \hline
\multirow{2}{*}{Experiments} & \multirow{2}{*}{Methods} & \multicolumn{3}{c|}{DTU testing set}  & \multicolumn{3}{c}{ETH3D training set}\\
 \cline{3-8}
  & & Acc. $\downarrow$ & Comp. $\downarrow$ & Overall $\downarrow$ & Acc. $\uparrow$ & Comp. $\uparrow$ & $F_1$-score $\uparrow$ \\
 \hline
 \multirow{4}{*}{Diffusion Models} & (\uppercase\expandafter{\romannumeral1}) w./o. diffusion & 0.324 & 0.312 & 0.318 & 72.68 & 70.21 & 70.69 \\
 & (\uppercase\expandafter{\romannumeral2}) noise in training & 0.339 & 0.320 & 0.329 & 73.55 & 72.48 & 72.47 \\
 & (\uppercase\expandafter{\romannumeral3}) noise in training \& testing & 0.328 & 0.304 & 0.316 & 73.06 & 72.08 & 72.01 \\
 & (\uppercase\expandafter{\romannumeral4}) \underline{w. diffusion} & \textbf{0.318} & \textbf{0.297} & \textbf{0.308} & \textbf{76.74} & \textbf{74.32} & \textbf{74.86} \\
 \hline
 \multirow{4}{*}{Diffusion Conditions} & (\uppercase\expandafter{\romannumeral5}) w./o. cost volume & 3.094 & 2.269 & 2.682 & 48.95 & 47.29 & 47.16 \\
 & (\uppercase\expandafter{\romannumeral6}) w./o. depth context & 0.333 & 0.298 & 0.316 & 44.94 & 43.54 & 43.07 \\
 & (\uppercase\expandafter{\romannumeral7}) w./o. image context & 0.320 & 0.302 & 0.311 & 72.46 & 72.89 & 72.08 \\
  & (\uppercase\expandafter{\romannumeral8}) \underline{w. all conditions} & \textbf{0.318} & \textbf{0.297} & \textbf{0.308} & \textbf{76.74} & \textbf{74.32} & \textbf{74.86} \\
 \hline
 \multirow{4}{*}{Diffusion Sampling} & (\uppercase\expandafter{\romannumeral9}) single sample & 0.355 & 0.324 & 0.340 & 65.96 & 69.15 & 66.77 \\
 & (\uppercase\expandafter{\romannumeral10}) w./o. confidence & 0.337 & 0.301 & 0.319 & 72.82 & 71.00 & 71.19 \\
 & (\uppercase\expandafter{\romannumeral11}) confidence for regularization & 0.335 & 0.301 & 0.318 & 75.44 & 72.20 & 73.22 \\
 & (\uppercase\expandafter{\romannumeral12}) \underline{w. confidence-based sampling} & \textbf{0.318} & \textbf{0.297} & \textbf{0.308} & \textbf{76.74} & \textbf{74.32} & \textbf{74.86} \\
 \hline
 \multirow{3}{*}{Diffusion Efficiency} & (\uppercase\expandafter{\romannumeral13}) single U-Net & 2.760 & 1.874 & 2.317 & 50.99 & 50.99 & 50.20 \\
 & (\uppercase\expandafter{\romannumeral14}) stacked U-Nets & 0.321 & 0.300 & 0.310 & 71.99 & 73.00 & 71.93 \\
 & (\uppercase\expandafter{\romannumeral15}) \underline{GRU} & \textbf{0.318} & \textbf{0.297} & \textbf{0.308} & \textbf{76.74} & \textbf{74.32} & \textbf{74.86} \\
 \hline

\end{tabular}
\label{tab:ablations}
\end{table*}

\customparagraph{Diffusion models} 
In this ablation, we remove the diffusion process from DiffMVS as the ablation model, named as \textit{DiffMVS0}. Specifically, we remove the timesteps and Gaussian noise throughout training and testing. %
However, we keep the condition encoder, 2D U-Net, convolutional GRU and confidence-based sampling for fair comparison. %
That is, \textit{DiffMVS0} is in fact a vanilla coarse-to-fine or iterative refinement approach. 
As shown in Row \uppercase\expandafter{\romannumeral1} of Table~\ref{tab:ablations}, \textit{DiffMVS0} performs worse on both DTU and ETH3D, with performance drops of 3.2\% and 5.6\%, respectively. 
This effectively demonstrates that the performance gains of DiffMVS stem from our designed diffusion mechanism. 
Since the diffusion process introduces random noise into the model, we experiment with introducing noise augmentation in \textit{DiffMVS0}. During training, we add random Gaussian noise on depth map and keep noise scale the same as DiffMVS for fair comparison. During testing, we try two settings: adding noise during testing (Row \uppercase\expandafter{\romannumeral3}) or not (Row \uppercase\expandafter{\romannumeral2}). 
We find that the zero-shot generalization ability on ETH3D improves when introducing random noise during training. However, the performance on both DTU and ETH3D are still worse than DiffMVS. 
Therefore, we conclude that the diffusion process is effective and improves the robustness of reconstruction.

\begin{figure}[t]
    \centering
    \includegraphics[width=\linewidth]{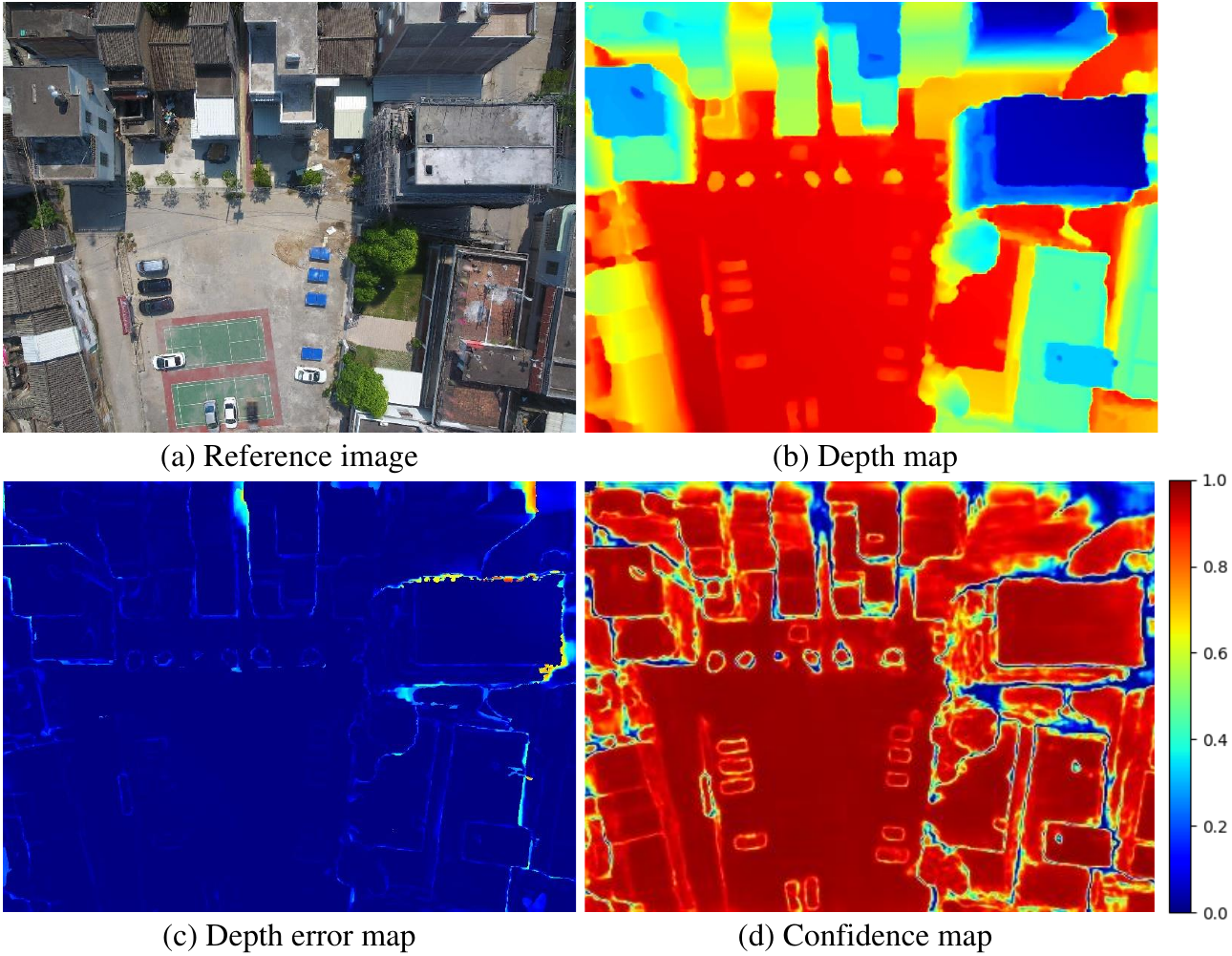}
    \caption{Visualization of reference image, depth map, depth error map and confidence map on the validation set of BlendedMVS~\cite{yao2020blendedmvs}.  %
    }
    \label{fig:conf_vis}
\end{figure}

\customparagraph{Diffusion conditions}
In this ablation, we investigate the efficacy of different diffusion conditions in our proposed condition encoder, including cost volume, depth context and image context features. As shown in Row \uppercase\expandafter{\romannumeral5} of Table~\ref{tab:ablations}, the reconstruction quality degrades a lot on both DTU and ETH3D without the cost volume as condition. This demonstrates that the cost volume plays a crucial role in our diffusion encoder as it encodes geometric matching information. 
By removing the depth context, the results in Row \uppercase\expandafter{\romannumeral6} of Table~\ref{tab:ablations} indicate that the depth context is beneficial to improve the generalization capability of our method on ETH3D. This is because the depth context features provide relative depth position information, allowing our condition encoder to consider corresponding depth ranges for new scenes. 
Furthermore, by removing the image context, we observe performance degradation from Row \uppercase\expandafter{\romannumeral7} of Table~\ref{tab:ablations}. In fact, the image context features provide semantic information of objects, reflecting depth continuities of objects to some extent. Therefore, the image context features can also facilitate our condition encoder.

\customparagraph{Diffusion sampling}
Recall that we propose confidence-based sampling strategy to adaptively adjust sampling range, where we generate multiple samples for the diffusion model. The confidence is learned in an unsupervised manner (Eq.~\ref{eq:loss_conf}) and used to linearly adjust the sampling range (Eq.~\ref{eq:conf_range}). 
In this ablation experiment, we first remove learned confidence and use a single sample, instead of generating multiple samples in a local range. As shown in Row \uppercase\expandafter{\romannumeral9} of Table~\ref{tab:ablations}, the performance on both datasets is the worst. 
Second, we remove learned confidence and generate multiple samples in a \textit{fixed} sampling range. As shown in Row \uppercase\expandafter{\romannumeral10} of Table~\ref{tab:ablations}, the performance improves when compared with using single sample (Row \uppercase\expandafter{\romannumeral9}). However, it is worse than our model (Row \uppercase\expandafter{\romannumeral12}). 
Third, we use learned confidence for regularization only (Eq.~\ref{eq:loss_conf}),~\ie, we do not use confidence to adjust the sampling range. We observe that the zero-shot generalization ability on ETH3D (Row \uppercase\expandafter{\romannumeral11}) is better than the ablation model without confidence (Row \uppercase\expandafter{\romannumeral10}). However, the performance on both datasets is still worse than our model since we further use confidence to adaptively adjust sampling range. 

In Fig.~\ref{fig:conf_vis}, we visualize the learned confidence on the validation set of BlendedMVS~\cite{yao2020blendedmvs}. Though we do not explicitly supervise the confidence during training, it reliably reflects the depth error distribution,~\ie, high confidence in regions with low depth error and low confidence in those with high depth error.

\customparagraph{Diffusion efficiency} 
In Sec.~\ref{sec:dif-ref}, we design a lightweight and effective diffusion network to denoise the depth residual, which takes advantage of U-Net and convolutional GRU. To further study its effectiveness, we replace this design by using a single U-Net and stacked U-Nets as the diffusion network, respectively. Note that for stacked U-Nets, we use $K=4$ U-Nets for fair comparison and each U-Net includes sampling, condition encoding and update. 
As shown in Row \uppercase\expandafter{\romannumeral13} and \uppercase\expandafter{\romannumeral14} of Table~\ref{tab:ablations}, the performance of single U-Net degrades significantly on both DTU and ETH3D, while the performance of stacked U-Nets drops slightly on DTU but obviously on ETH3D. Comparing Row \uppercase\expandafter{\romannumeral13} and \uppercase\expandafter{\romannumeral14} in Table ~\ref{tab:ablations}, we conjecture that it is important to use multiple updates in one diffusion sampling to improve convergence since only limited information is used in each update. The comparison between Row \uppercase\expandafter{\romannumeral14} and \uppercase\expandafter{\romannumeral15} in Table ~\ref{tab:ablations} verifies that the hidden state feature of the convolutional GRU can help the denoising process. Moreover, by introducing the convolutional GRU, the U-Net in our diffusion network can be reused to perform iterative refinement, instead of stacking multiple U-Nets. This effectively reduces model size with 33.7\% less parameters compared with stacked U-Nets, making our method more suitable for application scenarios with limited resources. 

\begin{table}[t]
\centering
\footnotesize
\caption{Evaluation of DiffMVS with different DDIM sampling steps $T_s$ on DTU. 
}
\begin{tabular}{c|cccc}
 \hline
 $T_s$ & Depth Error (mm) $\downarrow$ & Acc. $\downarrow$ & Comp. $\downarrow$ & Overall $\downarrow$\\
 \hline
 1 & 4.68 & \textbf{0.318} & \textbf{0.297} & \textbf{0.308} \\
 2 & \textbf{4.63} & 0.321 & \textbf{0.297} & 0.309 \\
 \hline
\end{tabular}

\label{tab:evaluation_steps}
\end{table}

\customparagraph{DDIM sampling} By default, we set DDIM sampling timestep as $T_s=1$. We change $T_s$ and summarize the results in Table~\ref{tab:evaluation_steps}. 
When we increase $T_s$, we find that depth accuracy slightly improves, while the quality of the point cloud is almost the same. Since we focus on both accuracy and efficiency, we set $T_s=1$ to reduce run-time. Unlike previous depth estimation works~\cite{ke2024repurposing} using diffusion models that initialize with pure random noise, our framework predicts an initial depth map $\bm{D}_{\text{init}}$ and then uses a diffusion model to refine it. Although not accurate enough, the initial depth map provides a relatively good initial value for the diffusion model to search for more accurate depth values in the neighborhood of $\bm{D}_{\text{init}}$. Therefore, our framework requires fewer sampling timesteps and thus reduces the run-time.

\begin{table}[t]
\centering
\footnotesize
\caption{Evaluation of CasDiffMVS with different noise scales on DTU and ETH3D. Note that on ETH3D, the noise scales become one-fourth of the original values because of our noise-scaling strategy.
}
\resizebox{\linewidth}{!}{
\begin{tabular}{c|ccc|ccc}
 \hline
\multirow{2}{*}{$(\sigma_2,\sigma_3)$} & \multicolumn{3}{c|}{DTU testing set}  & \multicolumn{3}{c}{ETH3D training set}\\
\cline{2-7}
& Acc. $\downarrow$ & Comp. $\downarrow$ & Overall $\downarrow$ & Acc. $\uparrow$ & Comp. $\uparrow$ & $F_1$-score $\uparrow$ \\
 \hline
 (0.25,0.05) & 0.361 & 0.323 & 0.342 & 70.01 & 74.94 & 71.73 \\
 (0.50,0.10) & \textbf{0.310} & \textbf{0.286} & \textbf{0.298} & \textbf{79.93} & 75.20 & \textbf{76.76} \\
 (1.00,0.20) & 0.315 & 0.296 & 0.306 &  76.95 & \textbf{76.20} & 76.09 \\
 \hline
\end{tabular}
}

\label{tab:evaluation_noise}
\end{table}

\customparagraph{Noise scale} To investigate the influence of noise scales in our diffusion model, we set different values for $(\sigma_2,\sigma_3)$ to train our CasDiffMVS and evaluate it on DTU test set and ETH3D training set. The results are reported in Table~\ref{tab:evaluation_noise}. We observe that our default setting $(0.50,0.10)$ achieves the best reconstruction performance on both sets. For the smaller one $(0.25,0.05)$, the random noise is too small to help our CasDiffMVS to avoid local minima and thus the performance degrades significantly. For the larger one $(1.00,0.20)$, the random noise is beneficial to our CasDiffMVS to escape local minima but the performance slightly drops. This is because our CasDiffMVS aims to refine a reasonable initial depth which does not contain too much noise. Too much noise will contaminate the initial depth and prevent our diffusion condition from generating favorable guidance. Therefore, our default setting can better reflect the noise level of the initial depth and introduce favorable perturbations to avoid local minima.

\begin{table}[t]
\centering
\footnotesize
\caption{Evaluation of DiffMVS with different random seeds on DTU. 
}
\begin{tabular}{ccc}
 \hline
 Acc.(mm) $\downarrow$ & Comp.(mm) $\downarrow$ & Overall(mm) $\downarrow$\\
 \hline
 0.3190 $\pm$ 0.0002 & 0.2976 $\pm$ 0.0006 & 0.3083 $\pm$ 0.0002 \\
 \hline
\end{tabular}

\label{tab:random_seed}
\end{table}

\customparagraph{Random seeds} We evaluate DiffMVS on DTU with 10 different random seeds and summarize the results (mean and standard variance) in Table~\ref{tab:random_seed}. We observe that the quantitative results are stable with different random seeds. 
In previous monocular depth estimation methods~\cite{ke2024repurposing} with diffusion models, the depth map is fully initialized using random noise, and thus specific modules are proposed to combat the diversity of diffusion models. In contrast, we use diffusion models to refine a reasonable initial depth map, $\bm{D}_{\text{init}}$, and search for more accurate depth values in the neighborhood of $\bm{D}_{\text{init}}$. Therefore, our framework is more stable w.r.t. random seeds.

\section{Conclusion}
In this paper, we introduce diffusion models in MVS for efficient and accurate reconstruction. 
We formulate depth refinement as a conditional diffusion process and propose a condition encoder for guidance. 
Moreover, we propose a confidence-based sampling strategy to adaptively adjust the per-pixel sampling range and thus improve accuracy. 
Instead of using large denoising U-Nets as classical diffusion models, we design a lightweight diffusion network, which combines a lightweight 2D U-Net and convolutional GRU, to improve both performance and efficiency. 
Based on our framework, we propose two novel MVS methods, DiffMVS and CasDiffMVS. 
Extensive experiments demonstrate that DiffMVS achieves competitive performance with state-of-the-art efficiency in both run-time and memory, while CasDiffMVS achieves SOTA performance on DTU, Tanks \& Temples and ETH3D. 
Because of the high efficiency, impressive performance and lightweight structure, our methods can serve as new strong baselines for future research in MVS.

\bibliographystyle{IEEEtran}
\bibliography{egbib}

\end{document}